\PassOptionsToPackage{table,dvipsnames}{xcolor}
\documentclass{muxlab}
\usepackage{amsmath,amssymb,mathtools,nicefrac}
\usepackage{algorithm}
\usepackage{algpseudocode}
\usepackage{amsmath,amssymb}
\newcommand{\sg}{\operatorname{sg}}
\newcommand{\LN}{\operatorname{LN}}

\renewcommand*{\backref}[1]{}
\renewcommand*{\backrefalt}[4]{\ifcase#1\relax\or(page #2)\else(pages #2)\fi}
\definecolor{BrandGreen}{HTML}{00754B}
\definecolor{BrandRed}{HTML}{B93430}
\definecolor{muxviolet}{HTML}{00754B}
\definecolor{muxlight}{HTML}{D7EBE2}
\definecolor{muxaccent}{HTML}{00754B}
\definecolor{muxdark}{HTML}{174A35}
\definecolor{muxrule}{HTML}{B4D4C3}
\definecolor{muxkeybox}{HTML}{F2F8F4}
\definecolor{muxlink}{HTML}{00754B}
\hypersetup{citecolor=BrandGreen,linkcolor=BrandGreen,urlcolor=BrandGreen}

\usepackage{wrapfig}
\usepackage{tikz}
\usetikzlibrary{calc,arrows.meta}
\definecolor{pgreen}{HTML}{C3E0C5}  % fills of Fig. 1
\definecolor{pred}{HTML}{F9BBBB}
\definecolor{dgreen}{HTML}{4E9F75}  % lines of Fig. 2
\definecolor{dred}{HTML}{D95F59}

\title{Tiny Recursive Message Passing}
\renewcommand{\authorlist}{%
 {\sffamily\color{muxfg}\textbf{Tommaso Salvatori}\textsuperscript{1,2},\;\;\textbf{Matteo Forasassi}\textsuperscript{1,2}}\\[2pt]%
 {\sffamily\color{muxfg}\textbf{Alexis Toumi}\textsuperscript{1},\;\;\textbf{Giovanni de Felice}\textsuperscript{1}}}
\renewcommand{\affiliationlist}{%
 {\small\textsuperscript{1}Relational Intelligence, Inc\quad\textsuperscript{2}Vienna University of Technology}\\[2pt]%
 {\small\texttt{\{name\}@rel-int.ai}}}
\abstract{
Message passing neural networks have shown remarkable performance in algorithmic and constraint reasoning, due to their ability to encode the structure of the task directly in the architecture. Recent work has, however, shown that better results can be achieved using recursive transformers trained with deep supervision and truncated recursion, which make extremely deep loops stable to train thanks to newly developed procedures such as adaptive halting and discrete deduction lattices. However, these models are tied to a fixed input size, and cannot generalize to inputs larger than those seen during training. In this work, we apply the same procedures to message passing networks, and show that graph structure and recursive training are complementary, leading to parameter-efficient models with strong zero-shot transfer to larger instances. First, we test our method on standard constraint satisfaction problems, where we reach $99\%$ test accuracy on Sudoku-Extreme with a model of $16$k parameters, $50\times$ fewer than the previous best; this model trains in about $10$ minutes on an H100, or in $9$ hours on a CPU. Second, we show that the graph structure allows zero-shot transfer to larger instances on mazes, N-Queens, and achieve state-of-the-art performance among neural methods on Boolean satisfiability.

\par\medskip
{\centering
  \small\sffamily
  \textcolor{BrandGreen}{%
    \textbf{Code:}\enspace
    \href{https://github.com/rel-int/TRMP}{%
      \textcolor{BrandGreen}{\nolinkurl{https://github.com/rel-int/TRMP}}}%
  }%
  \par

}}

\begin{document}
\maketitle

\section{Introduction}

Structured reasoning tasks consist of problems whose solution is computed by executing a specific set of instructions, or by searching over a known set of constraints. Such tasks are traditionally solved by symbolic methods, which are exact and scale to arbitrarily large instances, but require the solving procedure to be designed by hand. Neural networks instead learn this procedure directly from examples. To be competitive, however, they must exploit the structure of the problem and generalize to instances larger than those seen during training. Initially, message passing methods \citep{gilmer2017neural,scarselli2008graph} dominated the neural subfield of algorithmic reasoning, as they are able to wire the interaction patterns of the problem directly inside the structure of the model. This way, no capacity is spent learning which variables constrain which, and the symmetries of the problem hold by construction rather than being learned from  data \citep{xu2019can}. Furthermore, they can run any number of rounds, and thus spend more computation on harder instances \citep{selsam2018learning}. This is particularly important when it comes to generalizing to problems that are harder than the ones encountered during training. Learning, however, is performed using backpropagation through time (BPTT) on the unrolled computational graph, which is memory expensive and hence limits in practice how deep the recursion can go. Since a model trained through few rounds makes poor use of extra rounds at test time, this also limits how far it can be pushed on instances larger than those seen during training.

Recent work has attacked this bottleneck by iterating a recurrent transformer model many times while replacing full backpropagation through the trajectory with a cheaper gradient. Deep equilibrium models differentiate through the fixed point of the forward pass, giving a parameter update in $\mathcal{O}(1)$ memory \citep{bai19,bai2020multiscale}, while tiny recursive models use a coarser approximation, running the recursion detached and differentiating only its final segment \citep{jolicoeur2025less}. The final ingredient that makes such deep loops trainable is deep supervision, which keeps the gradient path short  \citep{graves2016adaptive,wang25hierarchical}. Together these have brought Sudoku-Extreme within reach of small models: the work introducing the benchmark reported $55\%$ accuracy with 27M parameters \citep{wang25hierarchical}, while models with 800k parameters now approach a perfect score \citep{davis2026lattice,sghaier2026probabilistic,huang2026equilibrium}. Whether these approximations carry over to message passing is not obvious: the gradient that detached recursion discards is exactly the one that travels along the edges, and different wirings of the same constraint satisfaction problem depend on it very differently. %We find that they do carry over, at a cost set by the propagation distance of the graph.

The transformer models that achieved the above results are parameter efficient even though they do not encode the structure of the problem. Can the parameter count of a recursive model be reduced even further by giving it the structure in the way message-passing architectures do, while keeping the inference and learning procedures that made looped transformers work? Can we use this modern machinery to design message passing architectures that generalize well to larger scale tasks at test time? We answer both questions positively: Deep supervision, detached recursion, and adaptive halting all presuppose a computation that decomposes into resumable segments carrying an explicit state. Recursive GNNs also carry such a state: at every round, each node reads the messages sitting on its edges together with its own memory, which is a tensor that can be paused and resumed like any other. What recursive GNNs lack is not this state but a training procedure built around it, which is what we provide. Moreover, since parameters are tied to vertex types rather than to vertices, a model trained on small instances can be run unchanged, and for more rounds, on larger ones. Architecturally, what we propose is close to recurrent relational networks \citep{palm2018recurrent}; our contribution is the training and inference procedure, together with an analysis of how it interacts with the choice of wiring. Our contributions are briefly as follows:
\begin{itemize}[leftmargin=*]
    \item We adapt the training and inference machinery of tiny recursive models to recursive GNNs. While deep supervision and truncated recursion carry over directly, adaptive halting and the deduction lattice require reformulation, namely per-vertex halting logits pooled with a smooth minimum, and a lattice defined over residual formulas for SAT.
    \item We show that encoding the graph structure and using the recursive machinery are complementary: On Sudoku-Extreme and maze-unique, a $32$k-parameter recursive lattice transformer and a recursive GNN with the same graph trained with BPTT are outperformed by our method, which combines the two.
    \item We show that the cost of truncating gradients grows with the propagation distance of the graph: dense wirings are robust to truncation, while factor graphs, the wiring that transfers across sizes, require longer gradient windows. We also show how a recent lattice-based method developed to improve performance of recursive transformers \citep{davis2026lattice}, when ported to message passing models, largely increases the robustness to truncated gradients.
    \item We conclude with experiments that test zero-shot generalization to larger instances: on N-Queens, we train on $10\times10$ boards and evaluate on boards up to $16\times16$; on Boolean satisfiability (SAT), we train on medium-size SR and 3-SAT formulas and evaluate on hard ones, where a single rollout solves 1.5× more formulas than NeuroSAT. Our models often match or outperform the current state of the art, despite using smaller models. We also test a recently introduced 4-SAT benchmark \citep{skenderi2026bench}, where we outperform all of the neural methods reported in the original work.  
\end{itemize}

\section{Related Works}

\paragraph{Recurrent message passing.}
The original graph neural network \citep{scarselli2008graph} was already recursive: a single transition function, shared across vertices and iterations, is iterated to the fixed point of a contraction. GGNNs \citep{li2015gated} drop the contraction, unroll a GRU for a fixed number of steps, and train by backpropagation through time. Implicit GNNs \citep{gu2020implicit} return to the equilibrium view. Because a GNN is a distributed algorithm whose number of rounds must grow with the problem \citep{loukas2019graph}, weight-tied networks can be run longer at test time to solve larger or harder instances \citep{tang2020towards,schwarzschild2021can,bansal2022end}. Belief propagation is the canonical recursive message-passing algorithm, extended to the neural case  with separate updates for variables and factors \citep{kuck2020belief,satorras2021neural,zhang2020factor}, which can be seen as a special case of relational and heterogeneous networks \citep{schlichtkrull2018modeling,wang2019heterogeneous}. Our modules also address neighbours individually, as in the port-numbering model \citep{angluin1980local,hella2012weak} and CPNGNN \citep{sato2019approximation}, and fall back on Deep Sets parameter sharing \citep{zaheer2017deep,maron2018invariant} when ports are interchangeable. %The closest precedents are Recurrent Relational Networks \citep{palm2018recurrent}, which solve Sudoku with deep supervision, and NeuroSAT \citep{selsam2018learning}, which passes messages over the bipartite structure of SAT and runs longer at inference than in training. %Both backpropagate through the full unrolled trajectory, which bounds their depth in practice.

\paragraph{Deep recursive computation.}
A complementary line of work has investigated how weight-shared models can be iterated to greater depths without backpropagating through the entire trajectory. Equilibrium GNNs allow message passing to approach arbitrary depth through fixed-point computation  \citep{yang2025implicit,georgiev2024deep}, while recurrent GNNs have explored adaptive message passing  at test time \citep{sun2026adaptive}. In parallel, recursive reasoning models have shown that simpler approximations, based on detached recursion, deep supervision, and adaptive halting, can train recurrent computations efficiently on reasoning problems \citep{jolicoeur2025less,wang25hierarchical}. In recent months, the performance of such recursive models has further improved thanks to novel inference procedures, which either inject Gaussian noise in the search  \citep{huang2026equilibrium,sghaier2026probabilistic}, or project the output into a lattice to guide search \citep{davis2026lattice}. % In this work, we implement, adapt, and test all of them to see which one performs better and in which case. 

\section{Background}
\label{sec:background}

\begin{figure}[t]
\centering
\includegraphics[width=0.8\linewidth]{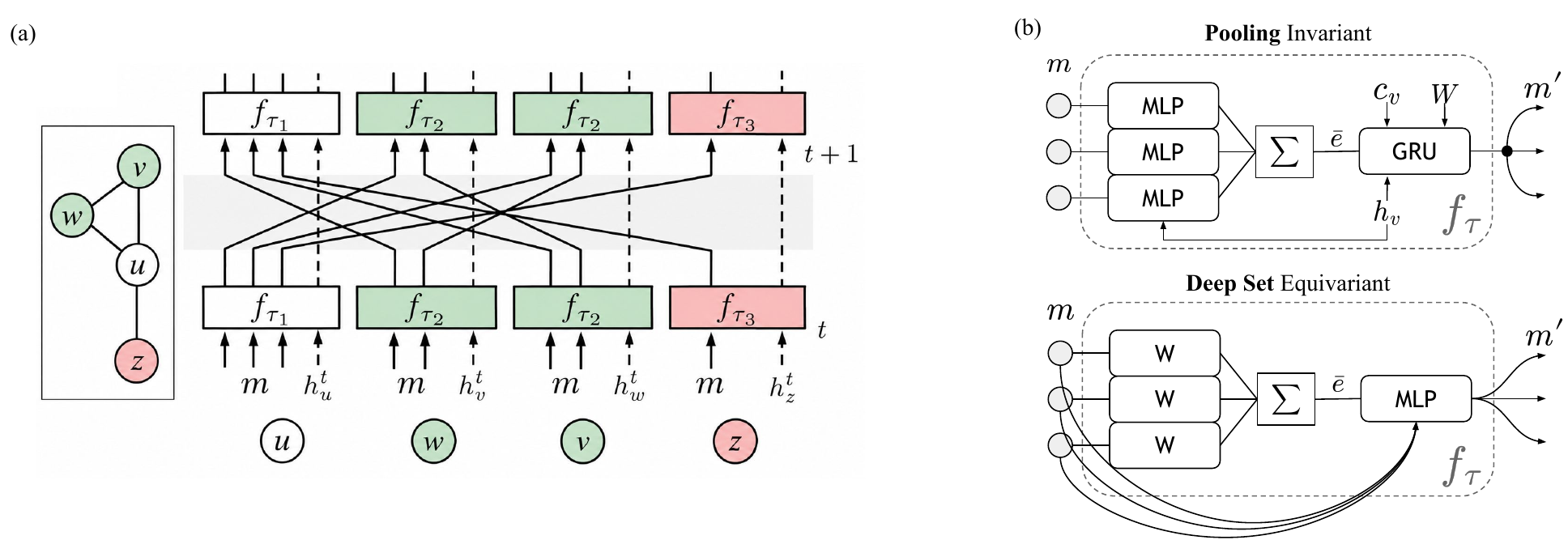}

\caption{(a) On the left, a graph with four nodes $(u,v,w,z)$, of  types $(\tau_1, \tau_2, \tau_2, \tau_3)$ respectively. On the right, the trajectory of a single message passing step on the corresponding model. The three types correspond to three different neural modules $f_{\tau}$, while modules of the same type are identical and share their weights. (b) Structure of the two main neural modules used in this work. The pooling module is invariant, and carries a latent state, used to compute the output when the iterative process has converged; the deep sets module is equivariant, and does not carry any latent memory.}
\label{fig:4x4}
\end{figure}

Let us assume we have a graph $(V,E)$, where every vertex $v$ is assigned a type $\tau$ from a finite set $\mathcal{T}$. In a message passing model, each vertex $v$ exchanges messages with its neighbours $w \in N(v)$: at every round $t$, the vertex reads the messages $m^{(t)}_{w\to v} \in \mathbb{R}^n$ it has received, combines them with its memory $h^{(t)}_v \in \mathbb{R}^{d_\tau}$, if it has one, and a fixed input $c_v \in \mathbb{R}^{d^c_\tau}$ to update the memory, and sends a new message $m^{(t+1)}_{v\to w}$ back to every neighbour. This message becomes the incoming message at that neighbour for the next round $t+1$. Repeating this for many rounds lets information travel across the whole graph. In neural message passing, all vertices of a given type $\tau$ share the same module $f_\tau$, whose memory, when present, is defined as the latent state of a gated recurrent unit (GRU) \citep{chung2014empirical}. For every vertex $v$ of type $\tau$, the module then performs the following operation:
\begin{equation}
\big( (m^{(t+1)}_{v\to w})_{w \in N(v)},\, h^{(t+1)}_v \big) = f_\tau\big( (m^{(t)}_{w\to v})_{w \in N(v)},\, h^{(t)}_v,\, c_v \big).
\label{eq:module}
\end{equation}
The input embedding $c_v$ (the clue) is how the datapoint $x_v$ enters the model, and it is the same at every round: For discrete inputs, such as the given digits of a Sudoku, $c_v = E x_v$.
The structure of each module is task dependent: we use larger modules for vertices that solve harder tasks, and design each module to respect the symmetries of its neighbours.  The two modules below, which cover almost every vertex type in this paper, are the special case in which all neighbours are interchangeable. When neighbours are not interchangeable, each edge is labelled with a port type, and the weights of the modules are shared only among edges of the same type. In what follows, $\mathrm{MLP}$ denotes a two-layer MLP (linear layer, ReLU activation, and a linear layer). All parameters are specific to the vertex type, and we omit the subscript $\tau$ for readability. % We also drop all bias terms from the notation for simplicity.

\paragraph{Pooling module.} This module is invariant to permutations of the neighbours, and is hence used when the neighbours of $v$ are interchangeable and $v$ needs only one, shared answer, as for the cells of a Sudoku. Every incoming message is encoded jointly with the memory by an $\mathrm{MLP}$, and the results are summed over all neighbours into $\bar{e}_v \in \mathbb{R}^{d_e}$, where $d_e$ is the embedding width of the module. This sum, together with the input, is passed to a GRU whose hidden state is the memory; we write $\mathrm{GRU}(\text{input}, \text{state})$. The new memory is then projected by a matrix $W_{\mathrm{out}} \in \mathbb{R}^{n \times d_\tau}$, and the same message is sent to every neighbour:
\begin{align*}
    e_{w\to v} &= \mathrm{MLP}([h^{(t)}_v, m^{(t)}_{w\to v}]), \\
    \bar{e}_v &= \textstyle\sum_{w \in N(v)} e_{w\to v}, \\
    h^{(t+1)}_v &= \mathrm{GRU}([\bar{e}_v, c_v], h^{(t)}_v), \\
    m^{(t+1)}_{v\to w} &= W_{\mathrm{out}}\, h^{(t+1)}_v.
\end{align*}
%
%Since $\bar{e}_v$ does not depend on the order of $N(v)$, reordering the neighbours changes neither $h_v'$ nor any outgoing message.

\paragraph{Deep Sets module.} This module is equivariant to permutations of the neighbours: reordering the neighbours reorders the outgoing messages correspondingly. It has no memory and no input, and is used when the neighbours are interchangeable but $v$ must answer each one individually, as for the rows, columns, and blocks of a Sudoku. Each incoming message is embedded by a matrix $W_{\mathrm{emb}} \in \mathbb{R}^{d_e \times n}$, shared across neighbours, and the embeddings are summed into a summary $\bar{e}_v$. For each neighbour $w$, an $\mathrm{MLP}$ then decodes the concatenation of its incoming message with this summary into the outgoing message \citep{zaheer2017deep}:
\begin{align*}
    e_{w\to v} &= \mathrm{ReLU}(W_{\mathrm{emb}}\, m^{(t)}_{w\to v}), \\
    \bar{e}_v &= \textstyle\sum_{w \in N(v)} e_{w\to v}, \\
    m^{(t+1)}_{v\to w} &= \mathrm{MLP}([m^{(t)}_{w\to v}, \bar{e}_v]).
\end{align*}
A complete forward pass hence works as follows: First, we embed each discrete input $x_v$ into $c_v$;  then, we unroll the message passing schema for a fixed number of rounds $T$, and compute the output of every vertex with memory via a linear map $\hat{y}_v = W_{\mathrm{y}}\, h_v^{(T)}$. During training, this output is used to compute a cross entropy loss, whose  gradients are computed via BPTT across all $T$ rounds. In what follows, we replace this pipeline with the machinery of modern recursive models, namely deep supervision, truncated recursion, and adaptive halting, so that the same modules can be trained without this cost. For more details, we refer to the Appendix.

\section{Methods}

We now discuss  (1) the division of training and inference in three different loops, the \emph{inner loop}, \emph{outer loop}, and \emph{deep supervision loop} \citep{wang25hierarchical}; (2) the use of truncated gradients, that only store a small number of iterations of the outer loop in memory \citep{jolicoeur2025less}; (3) an adaptive computation time (ACT) head, that is, a small linear layer that learns whether the model has converged or not, and is able to stop the deep-supervision loop early \citep{graves2016adaptive}; or, alternatively (4) the use of a discrete lattice that represents the intermediate inferred solution across steps \citep{davis2026lattice}. %Here, we intuitively sketch the methods; all the details can be found in the Appendix.
Such tecniques require the addition of a new latent state $a_v \in\mathbb R^{d_{\tau}}$, which we call \emph{answer state}, and a small cell in charge of updating it.
%: During the inner loop, the answer state $a_v$ is concatenated with $\bar e_v$ and $c_v$ as an additional input to the GRU of the pooling module. It is held fixed during the inner loop and updated by $\mathrm{GRU}_a$ at its end. 
This has been shown to be fundamental for propagating information across the different nested loops \citep{wang25hierarchical}.
%We have tested both the original formulation of the HRM, that uses an independent layer to perform the update of the answer state, and the TRM version, that uses the same module. Our early results showed that a smaller layer dedicated to the update of the answer state worked better, and hence we use this formulation throughout this work. 
We have also tested the addition of noise on a trained model as done in  probabilistic TRM, which allows one to compute multiple predictions from the same input by adding Gaussian noise at the end of every deep supervision step, and to pick the one that has the highest likelihood of being correct using the ACT head \citep{sghaier2026probabilistic}.

\paragraph{Looped Iterations}
We divide computation into three nested recurrent loops. The \emph{inner loop} consists of $R$ rounds as defined in Eq.~(\ref{eq:module}), during which messages $m$ and vertex memories $h$ evolve while the current answer state $a$, representing the current proposed solution of the network, is held fixed. At the end of the inner loop, the \emph{outer loop} updates $a$ and starts a new inner loop, so the complete recurrence can be written as
\begin{equation}
(m^{(k+1)},h^{(k+1)})=\Phi_\theta^{R}(m^{(k)},h^{(k)},c,a^{(k)}),
\qquad
a^{(k+1)}=\mathrm{LayerNorm}\big(\mathrm{GRU}_{a}(h^{(k+1)},a^{(k)})\big).
\label{eq:cycle}
\end{equation}
%
% \begin{equation}
% (x^{k+1},h^{k+1})=\Phi_\theta^{R}(x^k,h^k,u,a^k),
% \qquad
% a^{k+1}=\Phi_\theta^{a}(h^{k+1},a^k,u),
% \end{equation}
%
Here $\Phi_\theta^{R}$ denotes $R$ sequential applications of the shared vertex modules $f_\tau$ from Eq.~(\ref{eq:module}), with $c$ and $a$ held fixed, and $\theta$ collects the parameters. Superscripts $(k)$ index outer iterations rather than rounds: $h^{(k)}$ is the memory after $k$ inner loops, that is, after $kR$ rounds. In terms of size, the GRU cell for the outer loop is smaller than the main one, as it is always less than $10\%$ of the total parameters of the model. Increasing either the inner- or outer-loop depth therefore increases computation while leaving the model parameters and state dimensions unchanged. At the end of $K$ outer iterations, the answer vector of every cell is projected to the output space, $\hat{y}_v = W_{\mathrm{y}}\, a_v^{(K)}$, and a parameter update is performed. Instead of moving to the next batch, however, this whole process is repeated multiple times in a row, thus forming our third level of nested loop, called \emph{deep supervision}. We call each outer iteration a cycle: a deep-supervision step consists of $K$ cycles of $R$ rounds each, that is, $K \cdot R$ applications of Eq. (1), and we denote this configuration by $K \times R$ (e.g., $8 \times 12$ is 8 cycles of 12 rounds). %Implementation details are given in the Appendix.

\paragraph{Truncated gradients.}
As backpropagating through the entire trajectory would require storing all intermediate activations, we truncate the computational graph: earlier iterations are executed normally in the forward pass but their resulting state is detached before continuing, so gradients propagate only through a subset of the trajectory. If $K$ outer iterations are executed and gradients are retained only for the last $K_g \le K$, the computation is
\begin{equation}
s^{(K)} = G_\theta^{\,K_g}\!\left(\operatorname{sg}\!\left(G_\theta^{\,K-K_g}\big(s^{(0)}\big)\right)\right),
\end{equation}
where $s=(m,h,a)$, $G^j$ denotes $j$-fold composition, and $\operatorname{sg}$ denotes stop-gradient. Here, $U_\theta$ is the answer update in Eq.~(\ref{eq:cycle}), so $G_\theta=U_\theta\circ \Phi_\theta^{R}$ is one cycle. Truncation therefore changes neither the forward dynamics nor the parameters or latent states of the model; it only limits temporal credit assignment and reduces training memory from the full recursive trajectory to the subset for which gradients are stored. 

\paragraph{Adaptive computation time.}
To avoid always running the full supervision loop, we train a halting head that stops the training of a specific datapoint and moves to the following one \citep{jolicoeur2025less}. Unlike previous work, however, a linear head defined on the vertices of a graph produces one logit per vertex via $q_v = w_q\,\operatorname{sg}(a_v)$, and not one for the overall latent state. This creates the need to aggregate the results of the individual logits, in order to reach a global, graph-level decision. As a task is correct if and only if all of the individual cells are correctly classified, we pool with a smooth minimum $q=-\log\sum_v\exp(-q_v)$, so a single unconvinced vertex is enough to keep the whole graph from halting. To train such a module, we minimize a binary cross-entropy loss defined on every $q_v$, against whether the decoded value of its vertex $v$ is already correct.

Interestingly, such a halting head can also be used to compute the confidence of the model with respect to a specific solution. We hence use it to select the most promising candidate in a set of possible solutions: we run inference on a batch of $P$ identical instances of the same datapoint, and we perturb the answer state with Gaussian noise at the end of every deep supervision step, resulting in $P$ different predictions. Then, we use the halting head to select the solution with the highest confidence. In the original work, this has allowed a pre-trained TRM to go from $87\%$ to $98\%$ test accuracy on sudoku-extreme \citep{sghaier2026probabilistic}.

\paragraph{Discrete lattice.}
Deep-supervision steps are connected through a discrete lattice representing the partial solution accumulated so far \citep{davis2026lattice}. For each variable, the lattice distinguishes values that have already been committed from those that remain unresolved; hence a state $L_\ell$ represents not only a candidate solution, but also the information inferred up to deep-supervision step $\ell$. For example, in Sudoku tasks, the entry of each cell in $L_\ell$ is a multi-hot subset of $\{1,\dots,9\}$: a blank cell starts with all nine candidates alive, and successive projections can only kill candidates, never revive them, until a cell is left with one alive candidate (committed) or, in a failure case, none; the support of $L_\ell$ therefore encodes the full history of eliminations made up to step $\ell$, not merely a current best guess. After each inference phase, the model produces a distribution $p_\theta(\cdot\mid L_\ell)$ over possible values. A projection operator $\Pi$ converts this continuous prediction into discrete deductions. After each deep-supervision step, the projection $\Pi$ removes every non-given candidate whose predicted keep probability falls below a threshold, and declares the board in conflict if some cell is left with no candidate. If the board is neither solved nor in conflict, $\Pi$ then commits one randomly chosen undecided cell to a symbol sampled from the predicted distribution of the model over its surviving candidates.
The resulting lattice then becomes the problem presented to the next deep-supervision step, 
\begin{equation}
L_{\ell+1}=\Pi\left(p_\theta(\cdot\mid L_\ell),L_\ell\right).
\end{equation}
A partial solution that hits an internal conflict is reset and the process restarts from scratch, until a maximum number of deep supervision steps is completed. This creates two kinds of failures: \emph{timeout}, meaning that the model has failed to reach any solution after multiple trials, and \emph{wrong}, meaning that the model produces an actual, wrong prediction. In short, this process can be seen as a way of guiding the recursive model towards the correct solution by providing each deep supervision loop the discrete partial solution produced by the previous supervision step.

\section{Experiments}
\begin{wrapfigure}{r}{0.42\textwidth}
  \vspace{-1.5cm}
  \centering
  % ---------------------------------------------------------------
% Figure: constraints as hyperedges; the factor graph is the hypergraph
% (its incidence graph), the peer graph is its lossy 2-section.  Needs (preamble):
%   \usepackage{tikz} \usetikzlibrary{calc,arrows.meta}
%   \definecolor{pgreen}{HTML}{C3E0C5}  % fills of Fig. 1
%   \definecolor{pred}{HTML}{F9BBBB}
%   \definecolor{dgreen}{HTML}{4E9F75}  % lines of Fig. 2
%   \definecolor{dred}{HTML}{D95F59}
% ---------------------------------------------------------------
\begin{tikzpicture}[x=1cm, y=1cm, line cap=round, line join=round,
    var/.style={circle, draw, fill=white, line width=.45pt,
                inner sep=0pt, minimum size=2.2mm},
    fac/.style={rectangle, draw, line width=.45pt,
                inner sep=0pt, minimum size=2.1mm},
    lbl/.style={font=\scriptsize, anchor=base},
    peer/.style={black!65, line width=.5pt},
    inc/.style={black, line width=.5pt},
    queen/.pic={\fill[black!85] (-.075,-.055) -- (.075,-.055) -- (.085,.05)
                -- (.04,.005) -- (0,.075) -- (-.04,.005) -- (-.085,.05) -- cycle;}]

  % hyperedge blob: hull of the members, padded and outlined
  \newcommand{\hgEdge}[3]{% #1 fill, #2 outline, #3 path of members
    \begin{scope}[blend mode=multiply]
      \begin{scope}[transparency group]
        \draw[#2, line width=14.4pt, line join=round, line cap=round, fill=#2] #3;
        \draw[#1!70, line width=13pt, line join=round, line cap=round, fill=#1!70] #3;
      \end{scope}
    \end{scope}}
  % variable in both hyperedges: half green, half red
  \newcommand{\hgShared}[1]{%
    \fill[pgreen] (#1) ++(135:1.1mm) arc[start angle=135, end angle=315, radius=1.1mm] -- cycle;
    \fill[pred]   (#1) ++(-45:1.1mm) arc[start angle=-45, end angle=135, radius=1.1mm] -- cycle;
    \draw[line width=.45pt] (#1) circle[radius=1.1mm];}
  % split board cell (lower-left green, upper-right red), lower-left corner #1
  \newcommand{\hgSplit}[1]{%
    \fill[pgreen] (#1) -- ++(.4,0) -- ++(-.4,.4) -- cycle;
    \fill[pred]   (#1) ++(.4,0) -- ++(0,.4) -- ++(-.4,0) -- cycle;}

  % ============ top row: the factor graph IS the hypergraph ============
  % (i) hypergraph drawn as its factor graph: each hyperedge (blob) is a
  %     factor vertex (square) adjacent to exactly the variables it contains
  \begin{scope}[shift={(.6,0)}]
    \coordinate (a) at (-.86, .46);  \coordinate (c) at (.86, .46);
    \coordinate (b) at (-.86,-.46);  \coordinate (d) at (.86,-.46);
    \coordinate (v) at (0,0);
    \hgEdge{pgreen}{dgreen}{(a) -- (b) -- (v) -- cycle}
    \hgEdge{pred}{dred}{(c) -- (d) -- (v) -- cycle}
    \coordinate (f1) at (-.573,0);   \coordinate (f2) at (.573,0);
    \draw[inc] (f1) -- (a)  (f1) -- (b)  (f1) -- (v)
               (f2) -- (c)  (f2) -- (d)  (f2) -- (v);
    \node[fac, fill=pgreen] at (f1) {};
    \node[fac, fill=pred]   at (f2) {};
    \foreach \p in {a,b,c,d} \node[var] at (\p) {};
    \hgShared{v}
    \node[lbl] at (0,-.98) {hypergraph $=$ factor graph};
  \end{scope}
  % (ii) its 2-section, the peer graph: every hyperedge becomes a clique,
  %      and the edges no longer record which constraint they come from
  \draw[-{Stealth[length=1.6mm]}, line width=.5pt, black!70] (1.9,0) -- (3.56,0)
    node[midway, above=.4mm, font=\scriptsize, text=black] {2-section}
    node[midway, below=.6mm, font=\tiny, text=black!70, align=center] {forgets\\[-.3ex]hyperedges};
  \begin{scope}[shift={(4.5,0)}]
    \coordinate (a) at (-.68, .40);  \coordinate (c) at (.68, .40);
    \coordinate (b) at (-.68,-.40);  \coordinate (d) at (.68,-.40);
    \coordinate (v) at (0,0);
    \draw[peer] (a) -- (b) -- (v) -- cycle  (c) -- (d) -- (v) -- cycle;
    \foreach \p in {a,b,c,d} \node[var] at (\p) {};
    \hgShared{v}
    \node[lbl] at (0,-.98) {peer graph};
  \end{scope}

  % ================= bottom row: hyperedges in our tasks =================
  % --- Sudoku: a row (green) and a column (red)
  \begin{scope}[shift={(-.8,-3.05)}]
    \fill[pgreen] (0,.8) rectangle (1.6,1.2);
    \fill[pred]   (.4,0) rectangle (.8,1.6);
    \hgSplit{.4,.8}
    \draw[black!45, line width=.3pt] (0,0) grid[step=.4] (1.6,1.6);
    \draw[line width=.6pt] (0,0) rectangle (1.6,1.6) (.8,0) -- (.8,1.6) (0,.8) -- (1.6,.8);
    \foreach \r/\row in {0/{1,2,3,4}, 1/{3,4,1,2}, 2/{2,1,4,3}, 3/{4,3,2,1}}
      \foreach \d [count=\c from 0] in \row
        \node[font=\tiny, inner sep=0pt] at (.2+.4*\c, 1.4-.4*\r) {\d};
    \node[lbl] at (.8,-.32) {Sudoku};
  \end{scope}
  % --- N-Queens: a column (green) and an anti-diagonal (red)
  \begin{scope}[shift={(2.25-.8,-3.05)}]
    \foreach \r in {0,...,3} \foreach \c in {0,...,3} {
      \pgfmathtruncatemacro{\odd}{mod(\r+\c,2)}
      \ifnum\odd=1 \fill[black!9] (.4*\c,1.2-.4*\r) rectangle ++(.4,.4); \fi }
    \fill[pgreen] (.4,0) rectangle (.8,1.6);
    \foreach \r in {0,1,3} \fill[pred] (1.2-.4*\r,1.2-.4*\r) rectangle ++(.4,.4);
    \hgSplit{.4,.4}
    \draw[black!45, line width=.3pt] (0,0) grid[step=.4] (1.6,1.6);
    \draw[line width=.6pt] (0,0) rectangle (1.6,1.6);
    \foreach \r/\c in {0/1, 1/3, 2/0, 3/2} \pic at (.2+.4*\c, 1.4-.4*\r) {queen};
    \node[lbl] at (.8,-.32) {N-Queens};
  \end{scope}
  % --- SAT: two clauses sharing x3
  \begin{scope}[shift={(4.5,-2.25)}]
    \coordinate (x1) at (-.6, .34);  \coordinate (x2) at (-.6,-.34);
    \coordinate (x3) at (.05, 0);    \coordinate (x4) at (.68, 0);
    \hgEdge{pgreen}{dgreen}{(x1) -- (x2) -- (x3) -- cycle}
    \hgEdge{pred}{dred}{(x3) -- (x4)}
    \foreach \p in {x1,x2,x4} \node[var] at (\p) {};
    \hgShared{x3}
    \foreach \p/\t/\d in {x1/1/180, x2/2/180, x3/3/-100, x4/4/0}
      \node[font=\tiny, inner sep=0pt] at ($(\p)+(\d:.36)$) {$x_{\t}$};
    \node[font=\scriptsize, text=dgreen!85!black, anchor=base] at (-.22,.68) {$x_1{\lor}\lnot x_2{\lor} x_3$};
    \node[font=\scriptsize, text=dred!85!black, anchor=base]   at ( .52,-.70) {$\lnot x_3{\lor} x_4$};
    \node[lbl] at (0,-1.12) {SAT};
  \end{scope}
\end{tikzpicture}
  \caption{Top: two constraints (green, red) are hyperedges sharing a variable. Drawing each hyperedge as a vertex (square) adjacent to its variables gives the factor graph, the same object in bipartite form. Its 2-section, the peer graph, turns each hyperedge into a clique and forgets which constraint each edge comes from. Bottom: the same pattern in Sudoku (row, column), N-Queens (column, anti-diagonal), and SAT (two clauses).}
  \label{fig:hypergraph}
  \vspace{-1.2\baselineskip}
\end{wrapfigure}
Before applying message passing to a task, we must choose the graph that describes it, and hence the module each vertex runs. A constraint satisfaction problem is naturally a \emph{hypergraph}, whose hyperedges are the sets of variables that share a constraint: a row of a Sudoku, a diagonal of an N-Queens board, or a clause of a SAT formula (Fig.~\ref{fig:hypergraph}). The \emph{factor graph} is this hypergraph itself, drawn as a bipartite graph (its incidence graph): every hyperedge becomes a vertex adjacent to its variables, which communicate only through shared constraints. The \emph{peer graph} is instead derived from it, as its 2-section: two variables are adjacent whenever some hyperedge contains both, so every hyperedge becomes a clique. In a $4\times4$ Sudoku, the sixteen cells and twelve units (rows, columns, and blocks) give a peer graph with 56 edges, and a factor graph with 28 vertices and 48 edges.

\paragraph{Factor graphs and size transfer.} The parameters of a factor graph describe constraint types, not instances, so the same model runs on instances of any size. On boards, a cell keeps one neighbour per constraint at every size, three in Sudoku and four in N-Queens: what grows is only the arity of the constraints, which the Deep Sets modules pool over, and the number of rounds needed to propagate information, which a recursive model supplies by iterating longer. In the peer graph, instead, the peers of every cell grow with the board, and the constraints are lost: a ternary clause and the three binary clauses over its pairs yield the same triangle, and an all-different constraint becomes pairwise disequalities, which have the same solutions but, under local propagation, miss that a digit fits in only one cell of a unit. The advantage of the peer graph is distance: one round per cell-to-cell exchange rather than two, which makes it more robust to truncated gradients. We therefore compare the two wirings on Sudoku, and use factor graphs for size transfer.

\subsection{Sudoku-extreme and Maze-unique} 

\begin{table}[t]
\centering
\caption{Performance on Sudoku-Extreme and Maze-Unique.}
\vspace{-2ex}
\small
\setlength{\tabcolsep}{5pt}
\resizebox{0.90\textwidth}{!}{%
\begin{tabular}{l|ccc|cc|cccc}
\toprule
& \multicolumn{3}{c|}{Peer Graph no Lattice (32k)}
& \multicolumn{2}{c|}{Lattice (<24k)}
& \multicolumn{4}{c}{Baselines (32k)} \\
\cmidrule(lr){2-4}
\cmidrule(lr){5-6}
\cmidrule(lr){7-10}
Dataset
& Single Pass
& Noisy Single
& best-of-$64$
& Peer graph
& Factor graph
& LDT
& Graph-LDT
& RRN
& SE-RRN\\
\midrule

Sudoku-Extreme
& $98.7{\scriptstyle\,\pm\,0.06}$
& $98.2{\scriptstyle\,\pm\,1.2}$
& $\mathbf{99.9}{\scriptstyle\,\pm\,0.1}$
& $99.1{\scriptstyle\,\pm\,0.04}$
& $97.6{\scriptstyle\,\pm\,0.04}$
& $96.9{\scriptstyle\,\pm\,0.5}$
& $98.1{\scriptstyle\,\pm\,0.4}$
& $13.2{\scriptstyle\,\pm\,2.9}$
& $41.1{\scriptstyle\,\pm\,6.3}$\\

Maze-Unique
& $99.9{\scriptstyle\,\pm\,0.0}$
& $99.9{\scriptstyle\,\pm\,0.1}$
& $\mathbf{100}{\scriptstyle\,\pm\,0.0}$
& $99.9{\scriptstyle\,\pm\,0.0}$
& n.a.
& $18.6{\scriptstyle\,\pm\,3.2}$
& $99.4{\scriptstyle\,\pm\,0.2}$
& $5.8{\scriptstyle\,\pm\,2.7}$
& $11.3{\scriptstyle\,\pm\,4.2}$\\

\bottomrule
\end{tabular}%
}
\label{tab:main-results}
\end{table}

\begin{figure}[t]
\centering
\includegraphics[width=\linewidth]{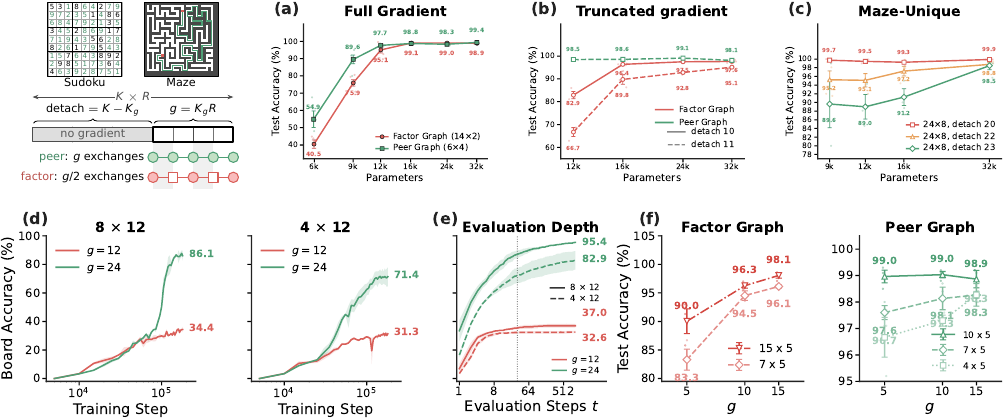}

\caption{ Top left: an example Sudoku  and maze. Below them, one
deep-supervision step of $K$ cycles of $R$ rounds: the first $K-K_g$ cycles
(\emph{detach} in the legends) run without gradient, and only the last
$g = K_g R$ rounds carry it. Within these $g$ rounds, information travels up to
$g$ cell-to-cell hops on the peer graph, but only $g/2$ on the factor graph. (a) Sudoku-extreme with the full gradient and lattice; (b) The same comparison under truncated backpropagation, $12 \times 6$;  (c) Maze-unique, 24  $\times$ 8, sweeping the truncation depth with the lattice; (d) Training curves when only $g$ rounds carry the gradients, no lattice; (e) Test-time extrapolation: board accuracy as a function of the number of inference steps $t$ for the same configurations; (f) Test accuracies using the lattice, under different numbers of cycles and rounds with gradient $g$. 
}
\label{fig:sudoku_maze}
\end{figure}

We test our method on the two constraint satisfaction puzzles most used in the recent literature: Sudoku-extreme \citep{wang25hierarchical}, boards of $9\times9$ cells to be completed, and Maze-unique \citep{huang2026equilibrium}, a variant of Maze-Hard restricted to $30\times30$ mazes with a single correct path between an entrance and an exit cell. We will investigate different questions: which one of the two graph structures performs better and survives the truncation of the gradients. We also compare the performance of a lattice-based model against that of the noisy halting head. We train models of different sizes, wirings, and truncation regimes on both tasks. A factor graph model with $16$k parameters trains on Sudoku-extreme in about ten minutes on a single H100, or in about nine hours on $32$ CPU threads of an Intel Xeon Platinum 8480C. Training on Maze-unique takes longer, since every round of message passing now runs over a grid of $900$ cells rather than a board of $81$. To select which configurations of cycles and rounds to report, we use the ones that have reached the highest validation accuracy over a single parametrization (average over $5$ seeds). To conclude, we evaluate the models without a lattice via a single pass with $128$ deep supervision steps, with a noisy pass with Gaussian noise of std $1.0$, via $64$ parallel noisy passes, which we call best-of-64.

\paragraph{Datasets.} Sudoku-extreme has $1000$ training, $300$ validation, and $1000$ test boards, selected as in \citep{davis2026lattice}. The symmetry group of Sudoku is the direct product of two groups: the group of geometric symmetries, defined by the wreath product $(S_3 \wr S_3) \wr S_2$, of order $3{,}359{,}232$, generated by permuting rows within a band, bands, columns within a stack, and stacks, and by transposition. The second is the group $S_9$ of digit relabelings. Since the geometric group is the automorphism group of both the peer and the factor graph, our models are equivariant to the group of geometric symmetries by construction. Hence, we augment only with random digit relabelings. Maze-unique has disjoint train and test splits of $1000$ mazes each, and a validation set of $300$ datapoints. As augmentation, we add the swapping of the entrance and exit points. To conclude, for every table we generate the peer graph such that pairs of connected walkable cells in the maze are connected in the graph.

\paragraph{Baselines.} To separate the contributions of architecture and training, we use four baselines; three transformer-based ones and a graph one. The first is a $32$k-parameter lattice deduction transformer (LDT) \citep{davis2026lattice}, which differs from our lattice models in its architecture. As this model does not have knowledge of the graph structure, we also implement a graph-LDT, where we embed the structure of the graph inside the architecture by masking connections that are not adjacent. Then, we use the symbol-equivariant (SE-RRN) recursive model \citep{freinschlag2026symbol}. As a message passing alternative, we consider RRNs \citep{palm2018recurrent} with the same shape and parameters as our models, so that they isolate the advantage given by the training procedure. We have tuned the hyperparameters of all baselines for the small-parameter regime. We have also tested our baselines when using models of $16k$ parameters, and observed a drop in performance, with none of them managed to get past $20\%$ board accuracy on Sudoku-extreme.  The performance, reported in Table~\ref{tab:main-results}, shows that our peer graph model outperforms all of the baselines at such small scales.

\subsubsection{Results}

With the full gradient, both wirings reach about 99\% on Sudoku-Extreme from 16k parameters onwards (Fig.~\ref{fig:sudoku_maze}(a)), and the peer graph reaches near-perfect accuracy on Maze-unique (Table~\ref{tab:main-results}). On a single-pass evaluation, the peer-graph lattice model outperforms the standard one on Sudoku-Extreme, and matches it on Maze-unique, even though it needs a larger amount of compute at test time, while the factor-graph lattice model stays slightly below both. Increasing the test-time compute of non-lattice based models and scoring them as done in probabilistic TRMs allows them to reach the best results, 99.9\% on Sudoku-Extreme and 100\% on Maze-unique with 64 passes. Under truncated gradients (Fig.~\ref{fig:sudoku_maze}(b)), the peer graph stays above 98\% at every size, while the factor graph degrades, especially for small models, and does not fully recover even with a longer gradient window. This is consistent with path length: in the bipartite factor graph, two cells of the same unit are two rounds apart rather than one, so a fixed truncation window covers half as many cell-to-cell exchanges. Maze-unique, whose graph has a far larger diameter, amplifies the effect (Fig.~\ref{fig:sudoku_maze}(c)): it needs more cycles and rounds, and exceeds 99\% at every size only with the longest gradient window. 

%The cost of truncating the gradient hence grows with the number of rounds information needs to cross the graph, and it is lowest for dense wirings such as the peer graph.

The accuracy results also hold when using the standard TRM recipe with the ACT head, and 32 deep supervision steps. We compare models with 32k parameters under one deterministic pass in Table~\ref{tab:main-results}, and the same model under a single and 64 noisy passes, with the best run selected by the halting head. Dropping the lattice slightly affects performance on Sudoku-Extreme in the deterministic setting even when running $1024$ deep supervision steps, and a single noisy pass does not improve on it, but it allows us to reach the best results once we use 64 parallel noisy passes. The lattice and the noisy halting head address the same difficulty from opposite directions: the lattice commits to partial deductions explicitly and restarts on conflict, while the halting head perturbs the answer state and picks among the resulting candidates.

\paragraph{Truncated gradients.} Let us call $g$ the number of rounds that carry gradients. In Figure~\ref{fig:sudoku_maze}, we study how the performance is affected when varying the number of cycles, gradient rounds $g$, and number of parameters of the model. Panel (d) varies the number of gradient-carrying rounds $g$ at fixed total depth, without the lattice. Accuracy rises sharply with $g$, more than doubling when going from $g=12$ to $g=24$, so a short window is not enough to learn Sudoku-Extreme from scratch. Figure~\ref{fig:sudoku_maze}(e) shows that this gap is further widened at inference: models trained with more gradients make better use of test-time iterations, while the others tend to converge earlier. This is what zero-shot transfer requires, since larger instances need more rounds than any seen during training. The credit horizon thus sets a ceiling at training time, so it should be treated as a per-task hyperparameter that grows with the propagation distance and diameter of the graph. We also observe a stark difference in performance between factor graphs and peer graphs (Fig.~\ref{fig:sudoku_maze}(f)), with the second being much more robust to the truncation of the gradients, due to its graph diameter being smaller; and between lattice and non-lattice methods (Fig.~\ref{fig:sudoku_maze}(d) and (f)): the lattice provides a simpler intermediate supervision signal, that allows the model to make improvements even using the small number of steps. This may not be the case for the standard supervision, as filling the next sudoku cell may need a larger number of backtracking steps than the ones that carry gradients, hence leaving the model with no useful signal. %Further discussions and experiments can be found in the supplementary material.

\subsubsection{Size Transfer}

\begin{wrapfigure}{r}{0.45\textwidth}
  \vspace{-1.2cm}
  \centering
  \includegraphics[width=\linewidth]{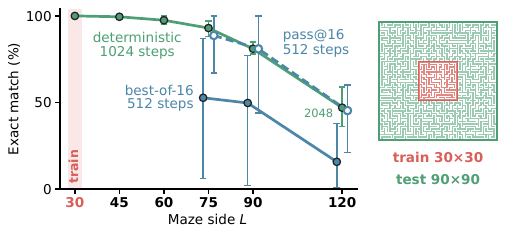}
  \vspace{-0.8cm}
  \caption{Zero-shot size transfer on mazes}
  \label{fig:maze-size}
  \vspace{-10pt}
\end{wrapfigure}

Since no parameter depends on the size of the graph, a model trained on $30\times30$ mazes runs unchanged on larger ones. We test this on  200 mazes of sizes $L\in\{45,\dots,120\}$ from the Maze-unique generator, using the ACT model with $32k$ parameters.

Run for a fixed budget, the model transfers far beyond its training size (Fig.~\ref{fig:maze-size}): with 2048 deep-supervision steps, it solves  $81\%$ of the mazes at $L=90$ and $47\%$ at $L = 120$, respectively $9$ and $16$ times the size of the maze it was initially trained. The weak component is the halting head, as solutions to puzzles stopped earlier largely underperform the deterministic solution. It also fails to select among samples: selecting the noisy rollout with the highest halting logit  loses to the deterministic pass at every size, while the pass@16 solution performs as well as the deterministic solver.  

\begin{figure}[t]
\centering

% ---------- Table ----------
\begin{minipage}[c]{0.60\textwidth}
\caption*{$^*$Used for model selection}
\vspace{-2ex}
\centering
\small
\setlength{\tabcolsep}{4pt}
\resizebox{\linewidth}{!}{%
\begin{tabular}{l|c|cccc}
\toprule
& \multicolumn{5}{c}{Test board size} \\
\cmidrule(lr){2-6}
Model & $10\times10^*$ & $12\times12$ & $14\times14$ & $15\times15$ & $16\times16$ \\
\midrule
RRN, $T=64$
& $0.83{\scriptstyle\,\pm\,0.21}$
& $0.68{\scriptstyle\,\pm\,0.15}$
& $0.41{\scriptstyle\,\pm\,0.11}$
& $0.14{\scriptstyle\,\pm\,0.10}$
& $0.02{\scriptstyle\,\pm\,0.03}$ \\

RRN, $T=128$
& $0.84{\scriptstyle\,\pm\,0.16}$
& $0.62{\scriptstyle\,\pm\,0.13}$
& $0.38{\scriptstyle\,\pm\,0.21}$
& $0.32{\scriptstyle\,\pm\,0.19}$
& $0.26{\scriptstyle\,\pm\,0.14}$ \\

\midrule
Ours, FG $8\times12$
& $0.96{\scriptstyle\,\pm\,0.03}$
& $0.98{\scriptstyle\,\pm\,0.01}$
& $0.55{\scriptstyle\,\pm\,0.27}$
& $0.38{\scriptstyle\,\pm\,0.31}$
& $0.28{\scriptstyle\,\pm\,0.26}$ \\

Ours, FG $10\times12$
& $0.97{\scriptstyle\,\pm\,0.02}$
& $0.88{\scriptstyle\,\pm\,0.14}$
& $0.63{\scriptstyle\,\pm\,0.44}$
& $0.61{\scriptstyle\,\pm\,0.44}$
& $0.57{\scriptstyle\,\pm\,0.34}$ \\

Ours, FG $12\times12$
& $0.99{\scriptstyle\,\pm\,0.01}$
& $\mathbf{0.98}{\scriptstyle\,\pm\,0.02}$
& $\mathbf{0.83}{\scriptstyle\,\pm\,0.12}$
& $\mathbf{0.67}{\scriptstyle\,\pm\,0.30}$
& $\mathbf{0.63}{\scriptstyle\,\pm\,0.35}$ \\
\bottomrule
\end{tabular}%
}
\end{minipage}
\hfill
% ---------- Figure ----------
\begin{minipage}[c]{0.30\textwidth}
\centering
\vspace{-0.2cm}
\includegraphics[width=\linewidth]{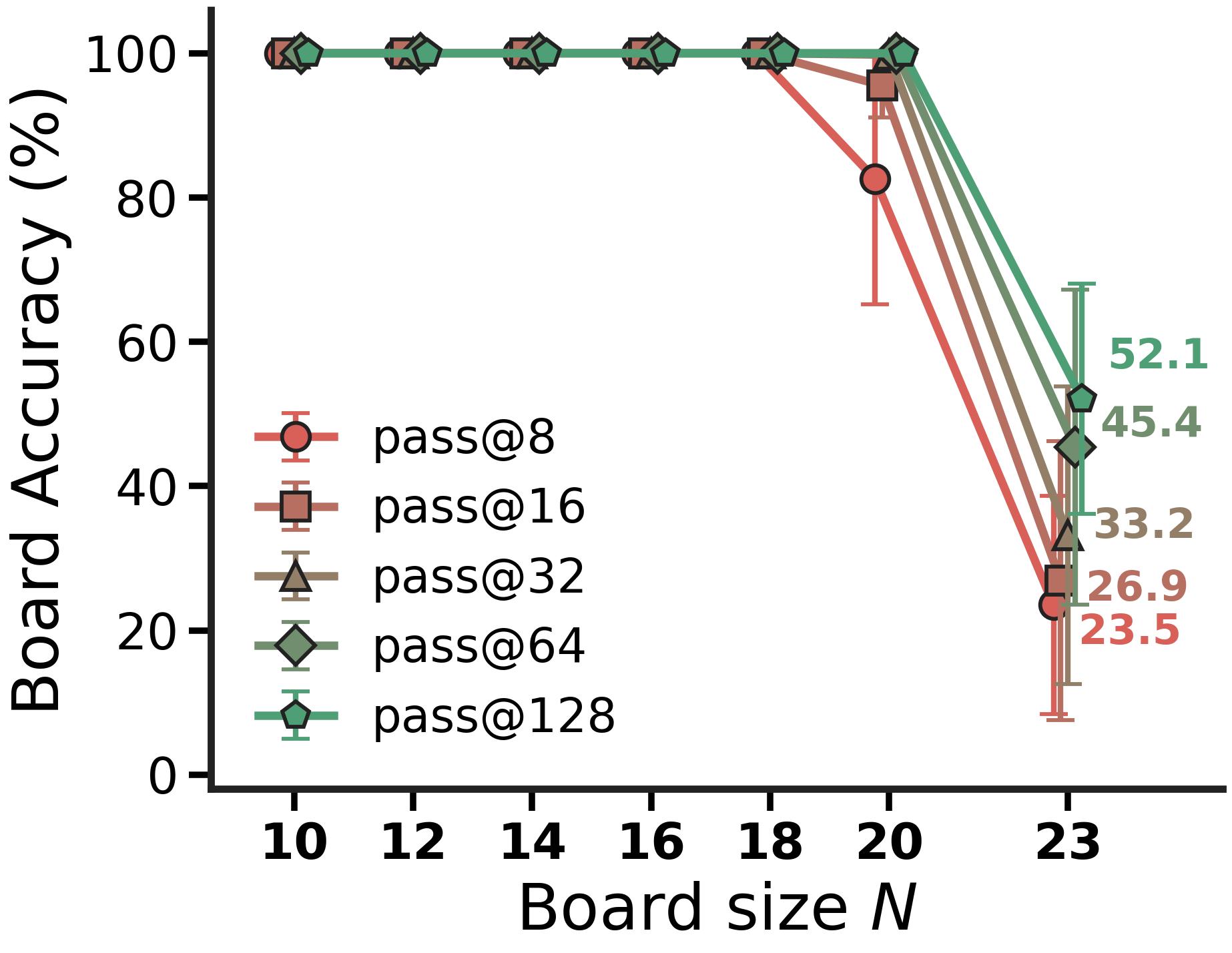}
\end{minipage}

\vspace{-0.2cm}
\caption{\textbf{Zero-shot size transfer on N-Queens.}
Left: Full board accuracy across test board sizes using a single deterministic pass.
Right: Full board accuracy via multiple probabilistic passes, assuming we have access to a verifier.}
\vspace{-0.4cm}
\label{fig:queens}

\end{figure}

\subsection{N-Queens}

To test generalization to boards larger than those seen in training, we use the N-Queens dataset \citep{nandwani2020neural}. Given an $N \times N$ chess board with some queens already placed, the task is to complete it to $N$ queens, no two sharing a row, column, or diagonal. For every complete $10\times10$ solution we sample $60$ partial boards, $20$ each with $1$, $2$, and $3$ queens. Since our models are equivariant to the eight symmetries of the square, we split by orbit under this group, so that no test board is a rotation or reflection of a training board. This gives 13,442 training and 2,372 validation boards. For larger boards $N \in \{12,\dots,16\}$, we sample $400$ solutions, keep $1$, $2$, or $3$ queens, giving $\approx 750$ test boards per size.

\paragraph{Model.} We use factor graph models, where every row, column, diagonal and anti-diagonal is a factor, and each cell of the factor graph is a node that receives an embedding of its clue (queen or empty) at every step. Every cell lies on exactly four lines, so it has exactly four ports for every $N$, while factors pool over their variable number of cells; no part of the model depends on $N$. As a baseline, we run an RRN on the same factor graph with the same widths, data, and target selection, trained as a single segment of $T \in \{ 64, 128\}$ rounds with BPTT and evaluated at a depth that increases with the board size, until we noted that more iterations were not improving the results anymore. We also tested $T = 256$, but the training with full BPTT did not fit into a single GPU. All models have $16$k parameters, and are trained with the ACT head. Since a partial board may have several completions, we use as label the board consistent solution with the lowest current cross-entropy, as done in previous work \citep{nandwani2020neural}. A board counts as solved if and only if it contains $N$ queens with no two sharing a line.

%Training follows the same setup described for Sudoku-extreme, with the difference that here we do not use the lattice, but only the training with the ACT head. As an incomplete board may have many valid completions, the target is selected from the set of solutions that currently has the lowest cross-entropy. This was done as picking a possible solution at random each step pushes toward the average instead of towards a specific solution. A predicted board counts as solved if it is a valid N-Queens configuration, with exactly $N$ queens and no two sharing a line, and every clue queen is preserved. Clues are clamped at decoding time, so the second condition holds by construction and correctness reduces to validity.

\paragraph{Results.}
All models transfer beyond their training size: On $16\times16$ boards, our best configuration solves $63\%$ on average (Figure~\ref{fig:queens} left). The RRN degrades faster than our models: $T=64$, drops to $2\%$ at $16\times16$, and only $T=128$ retains $26\%$, while requiring more memory and training time than our models. We have also tested the probabilistic formulation, where we have assumed we had access to a verifier that allows us to pick the correct solution out of sampled ones. The results in Figure~\ref{fig:queens}(right) show that our $12 \times 12$ configuration is able to reach near-perfect accuracy up to boards of size $20 \times 20$. We have observed that selecting the solution using the ACT head performs slightly worse, or on par with the deterministic single pass.

%The results, presented in Table~\ref{tab:queens}, show that the factor graph model is able to generalize well beyond its training regime, all the way to boards of size $16 \times 16$, although we have observed a very large variance across different seeds on larger boards. As expected, this was not the case for the peer graph, which gets near perfect accuracy on in-distribution test data, but collapses as soon as the board gets larger. The performance of the baselines is far inferior to our models, despite needing more memory and taking a longer time to train. This further confirms that the machinery recently developed by the recursive models is beneficial for message passing models as well. 

\subsection{Boolean satisfiability (SAT)}

We consider SAT, a standard testbed for algorithmic reasoning \citep{selsam2018learning, ozolins2022querysat, chang2025gatv2}. We focus on satisfying assignment prediction: A formula in conjunctive normal form (CNF)
is a conjunction of clauses, each a disjunction of literals, where a literal is a Boolean variable or its negation. Given a satisfiable CNF formula, the model must assign a truth value to every variable so that all clauses hold.

\paragraph{Datasets.} We use two benchmarks. From G4SATBench \citep{li2024g4satbenchbenchmarkingadvancingsat} we take
the random SR and 3-SAT families and its \emph{assignment} task, which keeps
only satisfiable formulas. Difficulty is the number of variables, drawn
uniformly per instance: $n\in[10,40]$ (easy), $[40,200]$ (medium), and
$[200,400]$ for SR or $[200,300]$ for 3-SAT (hard). Easy and medium provide
$80$k training, $10$k validation and test formulas each, and hard is test-only. RandCSPBench \citep{skenderi2026bench} instead mixes every
size $N\in\{16,\dots,256\}$ and every clause-to-variable ratio $\alpha$ into one
pool per family ($\alpha\in[3.0,5.0]$ for 3-SAT, $[8.0,10.0]$ for 4-SAT, in
steps of $0.1$), holding size fixed while sweeping across the satisfiability
threshold, where random instances are hardest. Its training pools contain $168$k
3-SAT and $80.8$k 4-SAT formulas, 
its test splits $42$k and $21$k. 
%We follow its score: instances solved, overthose any of its algorithms   solves, at $N\le 256$.

\paragraph{Model.} We use the VCG$^{*}$ encoding of G4SATBench: a bipartite
graph between variables and clauses with polarity-typed edges, i.e.\ the factor
graph of Sudoku and N-Queens with variables as cells and clauses as factors, the
two polarities entering as two port types. Variables run the pooling module and
emit one logit each; clauses run the Deep Sets module. No parameter depends on
the number of variables or on clause length, so one model runs on formulas of
any size. Since a formula admits many satisfying assignments, we train on the
unsupervised loss of G4SATBench, the negative log-probability that each clause
is satisfied. Without lattice, the ACT head scores clauses rather than
variables, pooled per instance by a size-corrected smooth minimum, and halts the
recursion once that score clears a threshold; best-of-$P$ perturbs the answer
state and keeps the rollout the head scores highest. With the lattice, every
variable is undecided, committed, or in conflict, and the loss is conditioned on
the commitments, so the model learns to complete a partial assignment.

\paragraph{Inference and baselines.} The lattice model reads the assignment off
the commitments where present and the readout elsewhere, then repairs it: it
picks an unsatisfied clause at random, commits one of its free variables so that
its literal becomes true, and restarts a chain whose commitments conflict. It
therefore never returns a wrong assignment, only timeouts. Methods also differ in
whether they evaluate the formula at inference. NeuroSAT, GGNN and our
no-lattice model do not: the latter halts and selects rollouts from a learned
score rather than from a satisfaction check. Our lattice model and QuerySAT
instead round the current assignment and test every clause as they run, which
steers the next step and lets them stop as soon as the formula is satisfied. As
baselines we use NeuroSAT and GGNN on G4SATBench, and NeuroSAT and QuerySAT on
RandCSPBench.

% As a formula admits many satisfying assignments, we do not supervise against a target,
% but minimize the unsupervised satisfiability loss of G4SATBench, i.e.\ the negative
% log-probability that each clause is satisfied. Without lattice, we train with the ACT
% head as for N-Queens; since no per-variable target exists, the head scores clauses
% rather than variables, and pools them with a smooth minimum corrected for the number of
% clauses. With lattice, each variable is either undecided, committed, or in conflict, and
% the loss is computed on the residual formula left by the committed variables. At
% inference, instead of the projection of Eq. (\ref{eq:projection}), we commit a variable from a randomly
% chosen unsatisfied clause so that its literal becomes true, restarting on conflict. As
% conflicts are checked exactly, the model never returns a wrong assignment, only
% timeouts.

\paragraph{Results.} With a single rollout, our model roughly matches NeuroSAT
on both benchmarks: in distribution on G4SATBench it is within $1.2$ points on
three of the four splits and $5$ points below on 3-SAT easy, and on RandCSPBench
it is $0.9$ points below on 3-SAT and $0.5$ above on 4-SAT. It already
outperforms GGNN on the medium splits, and under zero-shot transfer to hard
formulas it exceeds NeuroSAT by a factor of $1.5$ ($8.0$ vs.\ $5.2$ on SR,
$17.7$ vs.\ $11.3$ on 3-SAT). Selecting among noisy rollouts with the halting
head improves every column, and more so the harder the split: best-of-$64$ gains
about $20$ points on the G4SATBench medium splits and reaches $4$--$5\times$
NeuroSAT on hard, and on RandCSPBench it draws level with QuerySAT ($90.9$ vs.\
$92.4$ on 3-SAT, $67.8$ vs.\ $66.6$ on 4-SAT) at a fifth of its parameters. The
lattice is the strongest recipe in the aggregate on both benchmarks: it solves
over $99\%$ of the easy G4SATBench formulas, nearly doubles NeuroSAT on the
medium ones, and is the strongest neural method on both RandCSPBench families
($95.6$ and $72.9$), the only one above QuerySAT on 3-SAT. On the hard splits,
however, it falls behind the no-lattice model, whose best-of-$64$ noisy rollouts generalize better to larger formulas.

%The results are presented in Table~\ref{tab:SAT}. With a single rollout, our model without lattice performs on par with NeuroSAT and outperforms GGNN on the medium splits, while using between five and six times fewer parameters. Sampling multiple noisy rollouts and selecting one with the halting head consistently improves the solve rate, by up to 20 points on the medium splits with 64 rollouts. The lattice-based model is the best in distribution: it solves almost every easy instance, and nearly doubles the solve rate of NeuroSAT on the medium splits. In the zero-shot transfer to hard instances, all our models outperform NeuroSAT, with the best-of-64 model reaching a solve rate roughly four to five times higher. Here the lattice-based model falls behind the noisy halting head. Note that both the multiple rollouts and the lattice search use more computation at test time than the baselines, which perform a single forward pass.

\begin{table}[t]
\centering
\caption{Solve rate on G4SATBench \citep{li2024g4satbenchbenchmarkingadvancingsat}
and RandCSPBench \citep{skenderi2026bench}.
Zero-shot transfer models are trained on medium and evaluated on hard. All baseline solve rates are taken from the original work. NeuroSAT implementation is the same in both RandCSPBench and G4SATBench.}
\label{tab:sat}
\small
\setlength{\tabcolsep}{2.8pt}
\begin{tabular}{l r cc cc cc cc}
\toprule
& & \multicolumn{6}{c}{G4SATBench} & \multicolumn{2}{c}{RandCSPBench} \\
\cmidrule(lr){3-8} \cmidrule(lr){9-10}
& & \multicolumn{2}{c}{Easy} & \multicolumn{2}{c}{Medium}
& \multicolumn{2}{c}{Hard} & \multicolumn{2}{c}{$N \le 256$} \\
\cmidrule(lr){3-4} \cmidrule(lr){5-6} \cmidrule(lr){7-8} \cmidrule(lr){9-10}
Model & Params & SR & 3-SAT & SR & 3-SAT & SR & 3-SAT & 3-SAT & 4-SAT \\
\midrule
NeuroSAT & $430$k & $79.79$ & $80.59$ & $37.25$ & $41.61$ & $5.19$ & $11.34$
         & $84.48$ & $47.52$ \\
GGNN     & $445$k & $78.75$ & $76.42$ & $21.18$ & $25.68$ & --     & --
         & --      & --      \\
QuerySAT & $363$k & --      & --      & --      & --      & --     & --
         & $92.38$ & $66.57$ \\
\midrule
Ours, ACT ($P=1$) & $65$k
& $79.1_{\pm1.5}$ & $75.6_{\pm0.7}$ & $36.6_{\pm1.1}$ & $42.8_{\pm3.4}$
& $8.0_{\pm0.4}$  & $17.7_{\pm3.3}$
& $83.6_{\pm1.1}$ & $48.0_{\pm4.7}$ \\
Ours, ACT best-of-$8$ & $65$k
& $86.1_{\pm0.9}$ & $82.7_{\pm0.7}$ & $47.6_{\pm3.4}$ & $54.9_{\pm3.0}$
& $16.4_{\pm1.4}$ & $32.2_{\pm3.7}$
& $88.5_{\pm0.7}$ & $60.5_{\pm3.2}$ \\
Ours, ACT best-of-$64$ & $65$k
& $89.6_{\pm1.1}$ & $86.7_{\pm1.0}$ & $55.8_{\pm5.0}$ & $63.2_{\pm2.2}$
& $\mathbf{26.7}_{\pm3.8}$ & $\mathbf{44.8}_{\pm3.5}$
& $90.9_{\pm0.8}$ & $67.8_{\pm2.7}$ \\
\midrule
Ours, lattice & $65$k
& $\mathbf{99.7}_{\pm0.1}$ & $\mathbf{99.5}_{\pm0.3}$
& $\mathbf{72.5}_{\pm1.3}$ & $\mathbf{76.1}_{\pm1.6}$
& $12.3_{\pm1.4}$ & $29.5_{\pm4.7}$
& $\mathbf{95.6}_{\pm0.5}$ & $\mathbf{72.9}_{\pm2.1}$ \\
\bottomrule
\end{tabular}
\vspace{-1ex}
\end{table}

\section{Conclusion}

Our results show that the machinery of recent recursive transformers (deep supervision, truncated recursion, adaptive halting and deduction lattices) carries over to typed message passing, and that it is complementary to encoding the graph structure. Together they yield models with 16k–65k parameters that match or exceed neural baselines up to an order of magnitude larger. Truncation, however, becomes more costly as graph diameter grows, making wiring a trade-off between robustness to truncation and size transfer. A natural next step is to use such models as learned heuristics inside exact solvers, for problems with known constraint structure such as scheduling, configuration and verification-derived SAT.

%Our results suggest that the machinery behind recent recursive transformers, namely deep supervision, detached recursion and deduction lattices, is not specific to transformers. Applied to typed message passing, it lets models with tens of thousands of parameters match much larger ones. The savings are in parameters, not compute, and come mostly from giving the model the structure of the problem instead of making it learn it. At test time, the lattice and noisy rollouts attack the same difficulty from opposite sides: one commits to deductions and restarts on conflict, the other explores perturbed trajectories. The natural applications are problems with a known constraint structure where general-purpose solvers are exact but can be slow on a fixed distribution of instances, such as scheduling, configuration and SAT instances from verification. There, small recursive message-passing models, cheap to train and run, could serve as learned heuristics inside classical solvers: proposing assignments, guiding branching, or using the halting head to flag partial solutions worth trusting. We see this coupling of learned propagation and exact search as the most promising path from puzzles to practice.

\newpage

\bibliography{references}
\bibliographystyle{plainnat}

\appendix

% =====================================================================
%  Supplementary material: architecture and truncated-gradient experiments
%  on Sudoku-Extreme and Maze-unique.
%
%  Requires: amsmath, amssymb, booktabs.  Drop in after \appendix.
%  Literature references are written out in plain text and marked
%  "% CITE" so that they can be replaced by the paper's own bib keys.
%
%  Cross-references to the main text (rename to your own labels):  % REF
%    sec:background   Section 3 (Background)
%    sec:methods      Section 4 (Methods)
%    eq:main-module   Eq. (1), the vertex module f_tau
%    eq:main-loops    Eq. (2), inner/outer loop
%    eq:main-trunc    Eq. (3), truncated recursion
%    eq:main-commit   Eq. (4), lattice projection
%    eq:main-lattice  Eq. (5), lattice update
% =====================================================================

% =====================================================================
%  Supplementary material: architecture and truncated-gradient experiments
%  on Sudoku-Extreme and Maze-unique.
%
%  Requires: amsmath, amssymb, booktabs.  Drop in after \appendix.
%  Literature references are written out in plain text and marked
%  "% CITE" so that they can be replaced by the paper's own bib keys.
% =====================================================================

\newpage

\section{Pseudocode}

% Preamble:
% \usepackage{algorithm}
% \usepackage{algpseudocode}
% \usepackage{amsmath,amssymb}
%\newcommand{\sg}{\operatorname{sg}}
%\newcommand{\LN}{\operatorname{LN}}

\begin{algorithm}[t]
\caption{Training with deep supervision, truncated recursion and adaptive halting (one instance; batched in practice).}
\label{alg:train}
\begin{algorithmic}[1]
\Function{Cycle}{$s=(m,h,a),\ c$} \Comment{$G_\theta = U_\theta \circ F^R_\theta$}
  \For{$r = 1,\dots,R$} \Comment{inner loop, Eq.~(1); $a$ held fixed}
    \ForAll{$v \in V$ \textbf{in parallel}}
      \State $\big((m'_{v\to w})_{w\in N(v)},\, h'_v\big) \gets f_{\tau(v)}\big((m_{w\to v})_{w\in N(v)},\, h_v,\, c_v,\, a_v\big)$
    \EndFor
    \State $(m,h) \gets (m',h')$
  \EndFor
  \ForAll{$v \in V$ with memory} \Comment{answer update $U_\theta$, Eq.~(2)}
    \State $a_v \gets \LN\big(\mathrm{GRU}_a(h_v,\, a_v)\big)$
  \EndFor
  \State \Return $(m,h,a)$
\EndFunction
\Statex
\Require graph $(V,E)$, inputs $(x_v)$, labels $(y_v)$, configuration $K\times R$, gradient window $K_g \le K$, budget $D$
\State $c_v \gets E\,x_v$ for all $v \in V$
\State $s \gets s^{(0)}$ \Comment{initial messages, memories and answer states}
\For{$\ell = 1,\dots,D$} \Comment{deep-supervision loop}
  \For{$k = 1,\dots,K-K_g$} \Comment{no activations stored}
    \State $s \gets \Call{Cycle}{s,\, c}$
  \EndFor
  \State $s \gets \sg(s)$
  \For{$k = 1,\dots,K_g$} \Comment{gradients through the last $R_g = K_g R$ rounds, Eq.~(3)}
    \State $s \gets \Call{Cycle}{s,\, c}$
  \EndFor
  \State $\hat y_v \gets W_y\, a_v$, \quad $q_v \gets w_q^\top \sg(a_v)$ \quad for all $v$ with memory
  \State $\mathcal{L} \gets \sum_v \mathrm{CE}(\hat y_v,\, y_v) + \sum_v \mathrm{BCE}\big(q_v,\ \mathbb{1}[\arg\max \hat y_v = y_v]\big)$
  \State update $\theta$ with $\nabla_\theta \mathcal{L}$
  \State $s \gets \sg(s)$ \Comment{state carried to the next step, detached}
  \If{$-\log \sum_v e^{-q_v} > \delta_{\mathrm{halt}}$} \Comment{smooth-minimum halting}
    \State \textbf{break} \Comment{batch slot is refilled with a new instance}
  \EndIf
\EndFor
\end{algorithmic}
\end{algorithm}

\begin{algorithm}[H]
\caption{Inference. Best-of-$P$ with $P=1,\ \sigma=0$ is the deterministic single pass.}
\label{alg:inference}
\begin{algorithmic}[1]
\Function{BestOfP}{$x,\ P,\ \sigma,\ D$}
  \For{$i = 1,\dots,P$ \textbf{in parallel}}
    \State $c_v \gets E\,x_v$ for all $v$; \quad $s_i \gets s^{(0)}$
    \For{$\ell = 1,\dots,D$}
      \For{$k = 1,\dots,K$}
        \State $s_i \gets \Call{Cycle}{s_i,\, c}$
      \EndFor
      \If{$\ell < D$} \Comment{perturb the answer state}
        \State $a_{i,v} \gets a_{i,v} + \varepsilon_{i,v}$, \quad $\varepsilon_{i,v} \sim \mathcal{N}(0, \sigma^2 I)$
      \EndIf
    \EndFor
    \State $\hat y_{i,v} \gets \arg\max\, W_y\, a_{i,v}$, \quad $q_i \gets -\log \sum_v e^{-w_q^\top a_{i,v}}$
  \EndFor
  \State \Return $\hat y_{i^\star}$, \ with $i^\star = \arg\max_i q_i$ \Comment{selected by the halting head}
\EndFunction
\Statex
\Function{Lattice}{$x,\ D$}
  \State $L^{0} \gets$ lattice of $x$ \Comment{givens committed, blanks hold all candidates}
  \State $L \gets L^{0}$; \quad $s \gets s^{(0)}$
  \For{$\ell = 1,\dots,D$}
    \State $c_v \gets E\, L_v$ for all $v$ \Comment{multi-hot embedding of surviving candidates}
    \For{$k = 1,\dots,K$}
      \State $s \gets \Call{Cycle}{s,\, c}$
    \EndFor
    \State $p_v \gets \mathrm{softmax}(W_y\, a_v)$ for all $v$
    \State $(L,\ \mathit{status}) \gets \Call{Project}{p,\, L}$ \Comment{Eq.~(4)}
    \If{$\mathit{status} = \textsc{solved}$}
      \State \Return $L$
    \ElsIf{$\mathit{status} = \textsc{conflict}$}
      \State $L \gets L^{0}$; \quad $s \gets s^{(0)}$ \Comment{restart}
    \EndIf
  \EndFor
  \State \Return \textsc{timeout}
\EndFunction
\Statex
\Function{Project}{$p,\ L$}
  \ForAll{non-given $v$ and $j \in L_v$}
    \If{$p_v(j) < \delta$}
      \State $L_v \gets L_v \setminus \{j\}$ \Comment{candidates are only removed}
    \EndIf
  \EndFor
  \If{$L_v = \emptyset$ for some $v$}
    \State \Return $(L,\ \textsc{conflict})$
  \ElsIf{$|L_v| = 1$ for all $v$}
    \State \Return $(L,\ \textsc{solved})$
  \EndIf
  \State pick an undecided $v$ uniformly at random; sample $j \sim p_v$ restricted to $L_v$
  \State $L_v \gets \{j\}$ \Comment{commit one cell}
  \State \Return $(L,\ \textsc{continue})$
\EndFunction
\end{algorithmic}
\end{algorithm}

\newpage

\section{Recursive message passing with truncated gradients}
\label{app:trunc}

This appendix documents the models and protocols behind our Sudoku-Extreme and
Maze-unique experiments, in which back-propagation reaches only the final part
of the recursion. We first describe the recurrent message-passing network
shared by every model (\S\ref{app:trunc-arch}), then the two ways in which we
read it out and train it: a \emph{lattice} model that prunes sets of candidate
symbols and searches over them, and an \emph{ACT} model that decodes a one-hot
answer and halts adaptively (\S\ref{app:trunc-readouts}).
Section~\ref{app:trunc-grad} states precisely what the truncated gradient
computes. Sections~\ref{app:trunc-sudoku} and~\ref{app:trunc-maze} give, for
each task, the data, the hyperparameters and the evaluation protocol needed to
reproduce the reported numbers, together with the full results tables. We close
with the observations we found most instructive and
with practical notes on training (\S\ref{app:trunc-repro}).

\paragraph{Notation.}
Every vertex $v$ of a graph $(V,E)$, with neighbours $N(v)$, holds a memory
$h_v$, a clue embedding $c_v=Ex_v$ of its input $x_v$, and an answer state
$a_v$; every ordered pair of adjacent vertices carries a message
$m_{w\to v}$. Computation is organised in three nested loops: an
\emph{inner loop} of $R$ \emph{rounds} of message passing
(\S\ref{app:trunc-arch}), an \emph{outer loop} of $C$ \emph{cycles}, each
closed by one update of the answer state (Eq.~\eqref{eq:trunc-refresh}), and
\emph{deep-supervision steps}, each of which runs one full sweep of $C$ cycles
and is followed by a parameter update. We write the shape of a
deep-supervision step as $C\times R$. When training with truncated gradients
(\S\ref{app:trunc-grad}), the first $D$ cycles of a deep-supervision step
are executed without recording a computational graph, and back-propagation
reaches only the last $T=C-D$ cycles. The number of message-passing rounds that
receive credit,
\begin{equation}
  h \;=\; T R \;=\; (C-D)\,R ,
  \label{eq:trunc-h}
\end{equation}
is what we call the \emph{credit horizon}; it is the quantity varied by all of
our truncation studies. Setting $D=0$ recovers full back-propagation through
time, with $h=CR$. We write $\Sigma$ for the set of symbols a cell can take
and $k$ for the number of parallel chains or noisy rollouts used at test time.

% ---------------------------------------------------------------------
\subsection{Architecture}
\label{app:trunc-arch}

All models are recurrent message-passing networks on a fixed graph whose cell
vertices are the cells of the puzzle. A single module is shared by all
vertices of the same type, and is applied at every round, every cycle and every
deep-supervision step, so the parameter count does not depend on the amount of
computation ($C$, $R$), and the same weights can be iterated for as long as
desired at test time.

\paragraph{Graphs.}
For Sudoku we use two wirings of the same constraint structure. In the \emph{factor graph} (FG), the 81 cell vertices are joined to
27 unit vertices (nine rows, nine columns and nine $3\times3$ boxes); every
unit has nine neighbours and every cell has three, one per unit it belongs to.
In the \emph{peer graph} the unit vertices are removed and two cells are
adjacent whenever they share a unit, which gives every cell exactly 20
neighbours (eight in its row, eight in its column and four more in its box),
and 810 undirected edges in total. The peer graph is generated
programmatically from the factor graph, so the two wirings encode identical
constraints and differ only in how information is routed. In a round, every
vertex fires synchronously on the messages produced in the previous round, so a
cell-to-cell exchange takes one round in the peer graph and two rounds
(cell$\to$unit, unit$\to$cell) in the factor graph; the cell-to-cell diameter
is therefore 2 rounds in the peer graph and 4 in the factor graph.

For Maze-unique, each of the $30\times30=900$ cells has seven neighbours: its
four grid neighbours (a neighbour that falls outside the grid sends an
all-zero message, which is how a cell learns that it lies on the border), and
three unit vertices, namely its row, its column, and a single unit containing
the whole grid. There is a single wiring for this task.

No vertex carries a positional encoding, no edge carries a type, and all
vertices of the same type share their weights. As a consequence the networks
are \emph{exactly} equivariant to the symmetries of their graphs: for Sudoku,
to the geometric group generated by permutations of bands, of stacks, of rows
within a band, of columns within a stack, and by transposition; for the Maze,
to the eight symmetries of the square. We verify both properties numerically
(agreement to $10^{-12}$ in double precision), and neither task therefore needs
geometric data augmentation. Digit relabellings (the group $S_9$), by contrast,
are \emph{not} a symmetry of the architecture, because the input embedding and
the readout assign a meaning to each digit channel; we handle them by
augmentation (\S\ref{app:trunc-sudoku}).

\paragraph{Cell vertices: pooling module.}
Cells run a pooling module, which is invariant to the order of their
neighbours and broadcasts a single message to all of them. One round updates every cell $v$ synchronously as
\begin{align}
  e_w &= \operatorname{MLP}\big([h_v;\,m_{w\to v}]\big), \qquad w\in N(v),
         \label{eq:trunc-encode}\\
  h_v' &= \operatorname{LN}\!\Big(\operatorname{GRU}\big([\operatorname{mean}_{w}(e_w);\,c_v;\,a_v],\,h_v\big)\Big),
         \label{eq:trunc-gru}\\
  m_{v\to w}' &= W h_v' \qquad \forall\, w\in N(v),
         \label{eq:trunc-emit}
\end{align}
after which the new messages are routed along the edges of the graph. Here
$[\cdot\,;\cdot]$ denotes concatenation, $\operatorname{MLP}$ is a two-layer
perceptron with a ReLU, $\operatorname{GRU}(\cdot,h)$ is a standard gated
recurrent cell with hidden state $h$, and $\operatorname{LN}$ is layer
normalisation. We omit biases from the equations; they are present in the
models and counted in Eq.~\eqref{eq:trunc-params}. Every
incoming message is processed by the same MLP before the mean, which makes the
update invariant to the order of the neighbours, and a single message is
broadcast unchanged to all of them. Apart from the gates of the GRU there are
no residual connections, and no attention is used anywhere.

\paragraph{Unit vertices: Deep Sets module.}
The Sudoku factor-graph units and the Maze row, column and grid units run the
Deep Sets module, % CITE
which is equivariant to the order of their members. Given the messages $m_{w\to c}$ of the members $w\in N(c)$ of a unit $c$, the
unit replies to each member individually with
\begin{equation}
  \varphi_w = \operatorname{ReLU}\big(W_\varphi\, m_{w\to c}\big),\qquad
  m_{c\to w}' = \operatorname{MLP}\Big(\big[m_{w\to c};\,\textstyle\bigoplus_{w'\in N(c)}\varphi_{w'}\big]\Big),
  \label{eq:trunc-deepsets}
\end{equation}
which the member receives as one of its incoming messages. Units carry no
memory. The pooling $\bigoplus$ is a sum for Sudoku and a mean for the Maze,
where a unit has 30 (a row or a column) or 900 (the grid) members. All 27
Sudoku units share one module, so rows, columns and boxes are not
distinguished; the Maze uses one module for rows and columns and a second one
for the grid. The peer graph has no units at all: the message
$m_{w\to v}=Wh_w$ of a cell is read directly by its 20 neighbours.

\paragraph{Two time scales.}
After the $R$ rounds of a cycle, the answer state of every cell is refreshed
once by a second, smaller gated recurrent cell that reads the memory:
\begin{equation}
  a_v \leftarrow \operatorname{LN}_a\!\big(\operatorname{GRU}_a(h_v,\,a_v)\big).
  \label{eq:trunc-refresh}
\end{equation}
The memory $h$ is therefore a fast variable, updated at every round, while the
answer state $a$ is a slow variable, updated once per cycle; this mirrors the
latent and answer states of HRM and TRM.
Every readout is a linear function of $a$.

\paragraph{Initial state.}
A deep-supervision step starts from $h=0$, zero messages and $a=a_0$, where
$a_0$ is a learned vector initialised at zero. Because the initial state is
built outside the autograd graph, $a_0$ never receives a gradient and remains
zero in practice, so the starting state is a deterministic function of the
input alone. The lattice model re-initialises this state at every
deep-supervision step; the ACT model initialises it once per puzzle and carries
it across deep-supervision steps (\S\ref{app:trunc-readouts}).

\paragraph{Widths and parameter count.}
A model is specified by four widths $(n,m,n_{\mathrm{hid}},m_a)$: messages and
clue embeddings $m_{w\to v},u_v\in\mathbb{R}^{n}$, memories
$h_v\in\mathbb{R}^{m}$, the hidden layer of every MLP
$\mathbb{R}^{n_{\mathrm{hid}}}$, and answer states $a_v\in\mathbb{R}^{m_a}$.
Let $|\Sigma|$ be the number of input channels ($|\Sigma|=9$ for Sudoku,
$|\Sigma|=5$ for the Maze) and $n_{\mathrm{rel}}$ the number of Deep Sets
modules ($0$ for the peer graph, $1$ for the factor graph, $2$ for the Maze).
The number of trainable parameters is
\begin{align}
  P_{\mathrm{cell}} &= (m+n)\,n_{\mathrm{hid}} + n_{\mathrm{hid}}^2 + 2n_{\mathrm{hid}}
                      + 3m\,(n_{\mathrm{hid}}+n+m_a+m) + 8m + n\,m + n,\nonumber\\
  P_{\mathrm{rel}}  &= 3\,n\,n_{\mathrm{hid}} + n_{\mathrm{hid}}^2 + 2n_{\mathrm{hid}} + n,\nonumber\\
  P_{\mathrm{ans}}  &= 3m_a\,(m+m_a) + 8m_a,\nonumber\\
  P_{\mathrm{io}}   &= (|\Sigma|+1)\,n + 2(|\Sigma|+1)\,m_a + 2|\Sigma| + 1,\nonumber\\
  P &= P_{\mathrm{cell}} + n_{\mathrm{rel}}\,P_{\mathrm{rel}} + P_{\mathrm{ans}} + P_{\mathrm{io}},
  \label{eq:trunc-params}
\end{align}
where $P_{\mathrm{io}}$ collects the input embedding $E$, $a_0$ and the three
lattice readout heads of \S\ref{app:trunc-readouts}. The ACT models keep the
same heads and add an $(m_a+1)$-parameter halting head. The answer update
$P_{\mathrm{ans}}$ is below 10\% of the total for every model. To reach a
target budget we scan $n$ along the ray
$(n,m,n_{\mathrm{hid}},m_a)\approx n\cdot(1,\,2.615,\,5.231,\,1.308)$, build each
candidate, and keep the one whose exact parameter count is closest to the
budget. Table~\ref{tab:trunc-widths} lists the widths used in this appendix.

\begin{table}[t]
  \centering
  \small
  \caption{Widths $(n,m,n_{\mathrm{hid}},m_a)$ and exact parameter counts of the
  models in this appendix. Sizes are referred to by their nominal budget.}
  \label{tab:trunc-widths}
  \begin{tabular}{llcr}
    \toprule
    Task and wiring & Budget & Widths & Parameters \\
    \midrule
    Sudoku, peer graph   & 16k  & $(10,27,55,14)$  & 16{,}485 \\
    Sudoku, peer graph   & 24k  & $(12,34,65,16)$  & 23{,}978 \\
    Sudoku, peer graph   & 32k  & $(15,39,76,20)$  & 32{,}763 \\
    Sudoku, peer graph, with halting head & 32k & $(15,39,76,20)$ & 32{,}784 \\
    Sudoku, factor graph & 16k  & $(9,24,48,12)$   & 16{,}543 \\
    Sudoku, factor graph & 32k  & $(13,34,68,17)$  & 32{,}798 \\
    Maze                 & 32k  & $(12,31,64,16)$  & 34{,}786 \\
    \bottomrule
  \end{tabular}
\end{table}

% ---------------------------------------------------------------------
\subsection{Two readouts: the discrete lattice and adaptive halting}
\label{app:trunc-readouts}

The same network is used in two different systems, which differ in what is
carried from one deep-supervision step to the next, in how the output is read,
and in how the model is trained. We refer to them as the \emph{lattice} and the
\emph{ACT} models.

\subsubsection{Lattice model}

\paragraph{Carried object.}
The lattice model maintains, outside the recurrent state, a discrete lattice
$x_d\in\{0,1\}^{|V|\times|\Sigma|}$, where $d$ indexes deep-supervision steps and $V$ here denotes the cell
vertices: the row $x_{d,v}$ is the set of symbols still considered possible
for cell $v$. A given cell is a singleton; a blank Sudoku cell starts with all
nine digits, and a blank Maze cell with the two symbols \emph{free} and
\emph{path}. The input of a cell is its row of the lattice, so that
$u_v=E\,x_{d,v}$.

\paragraph{Readout and projection.}
Three linear heads read the answer state $a$: a keep probability
$p^{\mathrm{keep}}_{v,j}$ for every candidate $j$ of every cell, a symbol
distribution $p^{\mathrm{sym}}_v$ over $\Sigma$ for every cell, and a conflict
logit for every cell, whose log-sum-exp over cells gives a board-level conflict
probability $p^{\mathrm{conf}}$. Together they form the prediction
$p_\theta(\cdot\mid x_d)$. After each deep-supervision step, the non-differentiable projection $\Pi$
computes $x_{d+1}=\Pi\big(p_\theta(\cdot\mid x_d),x_d\big)$ as follows:
\begin{enumerate}
  \item every non-given candidate $j$ of cell $v$ with
        $p^{\mathrm{keep}}_{v,j}<\varpi=0.1$ is removed;
  \item the board is declared in conflict if some cell has no candidate left,
        or if $p^{\mathrm{conf}}>\varpi_{\mathrm{cls}}$
        ($0.5$ during training and $0.8$ when evaluating Sudoku);
  \item it is declared solved if every cell is a singleton and there is no
        conflict;
  \item otherwise one undecided cell is chosen uniformly at random and
        committed to a symbol sampled from $p^{\mathrm{sym}}_v$ at temperature
        $1.5$, restricted to its surviving candidates.
\end{enumerate}
Candidates are only ever removed, so the lattice descends monotonically; the
procedure is a DPLL-style solver whose propagation and branching heuristics are
learned. The recurrent state is rebuilt from the current lattice at every
deep-supervision step, so the only memory between steps is $x_d$ itself.

\paragraph{Training.}
Training uses a persistent pool of $B$ boards, which also serves as the batch.
At every optimisation step one deep-supervision step is run on every board in
the pool, the loss is computed, and the lattice is projected; boards leave the
pool and are replaced by fresh draws once they are finished, and after at most
100 steps. The loss is
\begin{equation}
  \mathcal{L} = \mathrm{BCE}_{w}\big(p^{\mathrm{keep}},\,x_d\odot x^\star_d\big)
  + 0.2\,\mathrm{CE}\big(p^{\mathrm{sym}}\big)
  + 0.1\,\mathrm{BCE}\big(p^{\mathrm{conf}}\big)
  \;\big[\,+\,0.1\,\mathrm{BCE}(\text{cell conflict})\,\big],
  \label{eq:trunc-lattice-loss}
\end{equation}
where $x^\star_d\in\{0,1\}^{|V|\times|\Sigma|}$ marks, for every cell, the
symbols that it takes in the solutions still compatible with $x_d$, and
$\mathrm{BCE}_w$ weights positives by $4$ and negatives by $0.5$. When a
commitment has excluded every solution, the board is labelled as a conflict.
The symbol cross-entropy is applied only to cells that are neither given nor
ambiguous in the target, and only on satisfiable boards; the per-cell conflict
term is used for the Maze only. Every differentiated cycle is read out, and the
terms are averaged over these readouts. For Sudoku, every fresh draw starts
from the raw puzzle with probability $0.20$, from a partial fill with the
correct digits with probability $0.55$, and from a partial fill in which
1--30\% of the filled cells are wrong with probability $0.25$, so that the
model learns both to complete assignments and to recognise contradictions; the
Maze uses an analogous mixture of partially solved and corrupted starts.

\paragraph{Inference by chain search.}
At test time each puzzle is attacked by $k$ independent stochastic chains run
in parallel. At every step every chain runs one deep-supervision step followed
by one projection; a chain that reaches a conflict restarts from the original
puzzle, and the first chain that reaches a solved board provides the answer,
which is graded afterwards (no oracle is involved). The compute budget is the
maximum number of deep-supervision steps per puzzle; a puzzle that is not
solved within the budget counts as a failure (a \emph{timeout}).

\subsubsection{ACT model}

\paragraph{Readout.}
The ACT model uses the symbol head alone and decodes every cell by
$\arg\max_j p^{\mathrm{sym}}_{v,j}$, writing the givens back; on the Maze the
$\arg\max$ is restricted to the symbols that the puzzle allows at that cell
(\emph{free} or \emph{path} on a blank cell), which reads the puzzle and never
the label. There is no lattice, no projection and no search, and the puzzle
encoding is never modified.

\paragraph{Halting.}
A linear head maps the answer state of every cell to a halting logit, and the
logits are pooled into a board-level logit by a smooth minimum:
\begin{equation}
  q_v = w_q\,\operatorname{sg}(a_v) + b_h,
  \qquad
  q = -\log\sum_{v}\exp(-q_v),
  \label{eq:trunc-halt}
\end{equation}
with $w_q$ initialised at zero and $b_h$ at $-5$, so that the model never halts
at initialisation. The smooth minimum $q$ lets a board stop only once every
cell is individually confident. Because the head reads $\operatorname{sg}(a)$,
halting is trained without shaping the representation, and its gradient
reaches only its own $m_a+1$ parameters.

\paragraph{Deep supervision with a carried state.}
A deep-supervision step runs $C$ cycles starting from the carried state
$(h,a)$; the state is passed on to the next step and detached in between, so no
gradient ever crosses the boundary between two deep-supervision steps.
Training follows the persistent-batch scheme of TRM: the batch consists of $B$
slots, every optimisation step advances every slot by one deep-supervision
step, and a slot is refilled with a fresh puzzle and a fresh state as soon as
$q>0$ or its trajectory has reached $n_{\mathrm{sup}}=32$ deep-supervision
steps. Slots are initialised at staggered depths (slot $i$ starts at phase
$i \bmod n_{\mathrm{sup}}$), which prevents the whole pool from expiring in the
same step, and the periodic gradient spikes that this otherwise causes. With
$p^{\mathrm{sym}}_{k}$ the readout after cycle $k$, the loss of a
deep-supervision step is
\begin{equation}
  \mathcal{L} = \frac{1}{C-D}\sum_{k=D+1}^{C}\mathrm{CE}\big(p^{\mathrm{sym}}_k,\,\text{target}\big)
  + \frac{1}{|V|}\sum_{v}\mathrm{BCE}\big(q_v,\,\mathbf{1}[v \text{ correct}]\big)
  + \lambda_{\mathrm{board}}\,\mathrm{BCE}\big(q,\,\mathbf{1}[\text{board correct}]\big),
  \label{eq:trunc-act-loss}
\end{equation}
with the cross-entropy averaged over cells and the halting targets computed
from the final readout of the step. The target is the single solution provided
with each puzzle.

\paragraph{Inference.}
We read a trained model in three ways: (i)~at \emph{fixed compute}, running a
prescribed number of deep-supervision steps and decoding the result;
(ii)~with \emph{halting}, stopping at the first deep-supervision step with
$q>0$, up to a cap; and (iii)~with \emph{best-of-$k$} selection, running $k$
rollouts in which Gaussian noise of scale $\sigma$ is added to $a$ at every
deep-supervision step, each rollout halting by its own rule, and returning the
rollout whose $q$ is highest. Best-of-$k$ uses only the model's own
confidence; we also report pass@$k$, the fraction of puzzles for which
\emph{any} rollout is correct, as an oracle upper bound on what selection could
achieve.

% ---------------------------------------------------------------------
\subsection{The truncated gradient}
\label{app:trunc-grad}

Write one cycle as $s_{k}=G_\theta(s_{k-1})$, where $s$ collects all memories,
messages and answer states, and $G_\theta$ applies $R$ rounds of
Eqs.~\eqref{eq:trunc-encode}--\eqref{eq:trunc-deepsets} followed by the answer
update~\eqref{eq:trunc-refresh}. Let $\ell_k$ be the loss of the readout after
cycle $k$. With $D$ detached cycles, the state reached by the differentiated
window is
\begin{equation}
  s_C = G_\theta^{\,T}\big(\operatorname{sg}\big[G_\theta^{\,D}(s_0)\big]\big),
  \qquad T=C-D,
  \label{eq:trunc-sg}
\end{equation}
where $\operatorname{sg}$ denotes stop-gradient, and training minimises
$\sum_{k=D+1}^{C}\ell_k$ while treating $s_D$ as a constant: every term of the
chain rule that would pass through $s_D$ into the first $D$ cycles is dropped.
Concretely, in both models:
\begin{itemize}
  \item the first $D$ cycles run inside a no-gradient context, so they build no
        graph, and their values are handed to the differentiated window as
        constants;
  \item the clue embedding $u=Ex$ is recomputed with gradient at the start of
        the differentiated window, so that the input embedding is still
        trained, through the last $T$ cycles only;
  \item every differentiated cycle is read out and supervised, while the
        detached cycles contribute no loss term at all;
  \item activation memory and backward cost scale with $h$ rather than with
        $CR$, while the forward computation is unchanged: at test time the
        network is identical to one trained with full gradients, and $D$ only
        affects how its parameters were obtained.
\end{itemize}
In the ACT model, deep-supervision steps are additionally separated by a
detach, so the credit horizon of a single optimisation step is $h$ in both
systems. For example, the peer-graph Sudoku recipe with $C\times R=12\times5$
and $D=11$ differentiates five rounds out of sixty, and the Maze lattice recipe
$24\times8$ with $D=20$ differentiates 32 rounds out of 192.

\paragraph{Stabilisation.}
Long credit horizons back-propagate through deep unrolled recurrences, whose
per-board gradient gain is heavy-tailed. All recipes clip the global gradient
norm. The ACT recipes additionally bound the influence of individual
trajectories: the gradient $g_b$ of board $b$ is rescaled by
$\min\!\big(1,\,\kappa\,\bar r/r_b\big)$, where $r_b=\lVert g_b\rVert/\ell_b$ is
the board's gradient per unit of its own loss, $\bar r$ is a slowly updated
reference for the median of $r_b$ (in the main recipes, an exponential moving
average with decay $0.99$ that moves by at most 1.4\% per accepted step), and
$\kappa=50$ in the main recipes and $5$ otherwise. An optimisation step whose
pre-clipping gradient norm exceeds a threshold is skipped altogether, since a
single very large gradient corrupts Adam's second-moment estimate for thousands
of steps. All reported models are the bias-corrected exponential moving average
(EMA) of the training weights (\S\ref{app:trunc-repro}).

% ---------------------------------------------------------------------
\subsection{Sudoku-Extreme}
\label{app:trunc-sudoku}

\paragraph{Data.}
We use the Sudoku-Extreme benchmark released with HRM, % CITE
downloaded from the Hugging Face dataset \texttt{sapientinc/sudoku-extreme}. A
deterministic build (seed 0) draws 1{,}000 training puzzles and a disjoint pool
of 18{,}000 validation puzzles from the official training split, and shuffles
the official test split, of which the first 1{,}000 puzzles form our test set.
Validation uses 500 puzzles from the validation pool. Every puzzle has a unique
solution.

\paragraph{Augmentation.}
Every time a training puzzle is drawn, its digits are relabelled by a fresh
uniformly random permutation in $S_9$, applied to puzzle and solution alike.
This is the only augmentation: the geometric symmetries are exact symmetries of
the network, and digit relabelling supplies the remaining factor of the
symmetry group, to which the network is not equivariant.

\paragraph{Lattice recipes.}
Table~\ref{tab:trunc-sudoku-lattice-hp} lists the lattice recipes. All use
AdamW with $\beta=(0.9,0.95)$ and weight decay on matrices only, a linear
warm-up followed by cosine decay to 5\% of the peak learning rate, bf16
autocast with TF32 matrix products. The width
study trains the peer graph at $12\times5$ with $D=11$ ($h=5$) at six budgets
between 6k and 32k parameters; its factor-graph counterpart uses the same shape
with $D=10$ ($h=10$, i.e.\ two differentiated cycles), because a single
differentiated factor-graph cycle is much less reliable.
The credit-horizon ablation fixes $R=5$ and the 16k models, and varies $C$ and
$D$ to obtain $h\in\{5,10\}$.

\begin{table}[t]
  \centering
  \small
  \caption{Lattice recipes for Sudoku-Extreme. $h$ counts rounds; a
  cell-to-cell exchange takes one round in the peer graph and two in the
  factor graph.}
  \label{tab:trunc-sudoku-lattice-hp}
  \begin{tabular}{lcccc}
    \toprule
    & Full gradient & Width study & \multicolumn{2}{c}{Credit-horizon ablation} \\
    \cmidrule(lr){2-2}\cmidrule(lr){3-3}\cmidrule(lr){4-5}
    & factor graph & peer graph & peer graph & factor graph \\
    \midrule
    Shape $C\times R$        & $14\times2$ & $12\times5$ & $\{4,7,10\}\times5$ & $\{7,10\}\times5$ \\
    Detached cycles $D$       & 0           & 11          & $C-1$, $C-2$        & $C-1$, $C-2$ \\
    Credit horizon $h$        & 28          & 5           & 5, 10               & 5, 10 \\
    Widths                    & $(9,24,48,12)$ & 6k--32k; 24k: $(12,34,65,16)$ & $(10,27,55,14)$ & $(9,24,48,12)$ \\
    Peak learning rate        & $6.8742\times10^{-3}$ & $4.8119\times10^{-3}$ & $6.8742\times10^{-3}$ & $6.8742\times10^{-3}$ \\
    Gradient-norm clip        & 1.0         & 0.5         & 1.0                 & 1.0 \\
    Weight decay              & 0.04821     & 0.04821     & 0.04821             & 0.04821 \\
    Warm-up fraction          & 0.1009      & 0.1009      & 0.1009              & 0.1009 \\
    Pool / batch size         & 512         & 512         & 512                 & 512 \\
    EMA decay                 & 0.9995      & 0.9995      & 0.9995              & 0.9995 \\
    Training steps            & 20{,}000    & 20{,}000    & 20{,}000            & 20{,}000 \\
    \bottomrule
  \end{tabular}
\end{table}

\paragraph{Lattice evaluation.}
From step 12{,}000, and then every 2{,}000 steps, the EMA weights are evaluated
on the 500 validation puzzles with 64 chains and a budget of 300
deep-supervision steps. The best validation checkpoint (ties broken in favour
of the later one) is then evaluated once on the 1{,}000 test puzzles, with 64
chains and a budget of 1{,}000 deep-supervision steps. For the full-gradient
factor-graph recipe, chains are checked for acceptance or reset every 4
deep-supervision steps on validation and at every step on test. We report exact
board accuracy: a board counts as solved only if all 81 cells are correct.

\paragraph{Lattice results.}
With full gradients, the 16k factor graph exceeds 99\%. Table~\ref{tab:trunc-sudoku-width} reports the truncated width study, with the
peer graph trained with full gradients at its best shape ($4\times6$, $D=0$) as
a reference. With a credit horizon of only five rounds out of sixty, the peer
graph matches full back-propagation between 12k and 16k parameters and exceeds
it at 24k, where it reaches 99.1\%. At 32k this recipe, tuned at 24k, falls
behind (98.1\% against 99.4\%). Truncation does not move the small-model cliff:
both regimes drop to about 54\% at 6k parameters. The factor graph trained on
the same shape, even with twice the credit horizon, loses at every width, and
dramatically so at 12k (82.9\% against 98.5\%).
Table~\ref{tab:trunc-sudoku-h} isolates the credit horizon at 16k parameters:
doubling $h$ from five to ten rounds helps the peer graph only marginally, but
lifts the factor graph by more than ten points.

\begin{table}[t]
  \centering
  \small
  \caption{Sudoku-Extreme, lattice model, truncated width study. Exact test
  accuracy (\%) on 1{,}000 puzzles, mean $\pm$ standard deviation over
  independent runs. The last row is the full-gradient reference.}
  \label{tab:trunc-sudoku-width}
  \begin{tabular}{lcccccc}
    \toprule
    & 6k & 9k & 12k & 16k & 24k & 32k \\
    \midrule
    Peer, $12\times5$, $D=11$ ($h=5$)
      & $53.9\pm6.5$ & $87.4\pm2.7$ & $98.5\pm0.2$ & $98.6\pm0.3$ & $\mathbf{99.1\pm0.2}$ & $98.1\pm0.4$ \\
    FG, $12\times5$, $D=10$ ($h=10$)
      & -- & -- & $82.9\pm2.2$ & -- & -- & $97.6$ \\
    \midrule
    Peer, $4\times6$, $D=0$ (full)
      & $54.9\pm10.8$ & $89.6\pm5.7$ & $97.7\pm0.9$ & $98.8\pm0.3$ & $98.3\pm0.6$ & $99.4\pm0.1$ \\
    \bottomrule
  \end{tabular}
\end{table}

\begin{table}[t]
  \centering
  \small
  \caption{Sudoku-Extreme, lattice model, credit-horizon ablation at 16k
  parameters with $R=5$. Mean exact test accuracy (\%) over all shapes and
  runs of each cell.}
  \label{tab:trunc-sudoku-h}
  \begin{tabular}{lcc}
    \toprule
    & $h=5$ (one cycle) & $h=10$ (two cycles) \\
    \midrule
    Peer graph   & 97.5 & 98.3 \\
    Factor graph & 85.3 & 95.7 \\
    \bottomrule
  \end{tabular}
\end{table}

\paragraph{ACT recipes.}
Table~\ref{tab:trunc-sudoku-act-hp} lists the ACT recipes, all on the 32k peer
graph. Our main recipe runs deep-supervision steps of $8\times12$ with $D=6$,
so that two of the eight cycles, i.e.\ $h=24$ of 96 rounds, are
differentiated. The credit-horizon ladder fixes $C\times R=9\times3$ and varies
only $D$, giving $h\in\{3,9,18,27\}$, the last being full back-propagation. The
truncation ladder fixes $R=12$ and compares two values of $C$ at
$h\in\{12,24\}$. All recipes use AdamW with $\beta=(0.9,0.95)$, weight decay
$0.04821$, 10\% warm-up and cosine decay to 5\%, EMA decay $0.9995$, a global
gradient-norm clip of $0.5$, 512 slots and $n_{\mathrm{sup}}=32$, and run in
single precision.

\begin{table}[t]
  \centering
  \small
  \caption{ACT recipes for Sudoku-Extreme, 32k peer graph, $(15,39,76,20)$.}
  \label{tab:trunc-sudoku-act-hp}
  \begin{tabular}{lccc}
    \toprule
    & Main recipe & Credit-horizon ladder & Truncation ladder \\
    \midrule
    Shape $C\times R$        & $8\times12$ & $9\times3$ & $8\times12$, $4\times12$ \\
    Detached cycles $D$      & 6 & 8, 6, 3, 0 & $C-2$, $C-1$ \\
    Credit horizon $h$       & 24 & 3, 9, 18, 27 & 24, 12 \\
    Peak learning rate       & $7.5\times10^{-4}$ & $1.5\times10^{-3}$ & $1.5\times10^{-3}$ \\
    Training steps           & 275{,}000 & 150{,}000 & 180{,}000 \\
    Board-level halting loss & $\lambda_{\mathrm{board}}=1$ & -- & -- \\
    Per-trajectory clip      & $\kappa=50$ & $\kappa=5$ & $\kappa=5$ \\
    Step rejection           & norm $>50$ & $>200\times$ running median & $>200\times$ running median \\
    Checkpoint selection     & \multicolumn{3}{c}{fixed compute, 128 deep-supervision steps, 500 validation puzzles} \\
    \bottomrule
  \end{tabular}
\end{table}

\paragraph{ACT evaluation.}
During training the EMA weights are evaluated on 500 validation puzzles, and
the checkpoint with the best fixed-compute board accuracy after 128
deep-supervision steps is kept. The final model is tested on the same 1{,}000
test puzzles as the lattice model, at fixed compute (32 to 2{,}048
deep-supervision steps), with halting at $q>0$, and with best-of-$k$ selection
over noisy rollouts; the noise scale $\sigma$ is chosen per run from a sweep,
and the selected values were 0.1 and 0.5.

\paragraph{ACT results.}
The main recipe reaches a mean exact test accuracy of 98.7\% with
deterministic halting (capped at 1{,}024 deep-supervision steps), and 99.9\%
with best-of-64 selection, matching the pass@64 oracle:
when the correct answer is among the rollouts, the halting head finds it.
Table~\ref{tab:trunc-sudoku-act-res} shows the two ladders. Credit sets the
level that training can reach: with $h=3$ or $h=9$ the model never leaves an
early plateau within 150{,}000 steps, $h=18$ escapes it after about 64{,}000
steps, and full back-propagation ($h=27$) after about 40{,}000. At equal credit
horizon the longer deep-supervision step is markedly better (89.6\% against
61.2\% at $h=24$), so the detached cycles are not wasted computation: they
bring the state to where the differentiated window can use its credit.

\begin{table}[t]
  \centering
  \small
  \caption{Sudoku-Extreme, ACT model: validation board accuracy (\%) at a fixed
  compute of 128 deep-supervision steps. Left: credit-horizon
  ladder after 150{,}000 steps. Right: truncation ladder after 180{,}000 steps.}
  \label{tab:trunc-sudoku-act-res}
  \begin{tabular}{cccc@{\hspace{2.5em}}cccc}
    \toprule
    \multicolumn{4}{c}{$9\times3$} & \multicolumn{4}{c}{$R=12$} \\
    \cmidrule(r{1.5em}){1-4}\cmidrule(l){5-8}
    $D$ & $h$ & Accuracy & Escape step & $C\times R$ & $D$ & $h$ & Accuracy \\
    \midrule
    8 & 3  & 15.8 & never      & $8\times12$ & 7 & 12 & 34.4 \\
    6 & 9  & 25.0 & never      & $8\times12$ & 6 & 24 & 89.6 \\
    3 & 18 & 41.4 & $\sim$64k  & $4\times12$ & 3 & 12 & 31.9 \\
    0 & 27 & 87.2 & $\sim$40k  & $4\times12$ & 2 & 24 & 61.2 \\
    \bottomrule
  \end{tabular}
\end{table}

Longer credit horizons are also harder to train. At a peak learning rate of
$1.5\times10^{-3}$ the $8\times12$, $h=24$ arm diverged, and the value in
Table~\ref{tab:trunc-sudoku-act-res} was obtained at $7.5\times10^{-4}$, the
learning rate of the main recipe. The $4\times12$, $h=24$ arm was also
unstable at $1.5\times10^{-3}$. Warm-started runs with credit horizons of 36 and
54 rounds diverged at the default learning rate as well (gradient norms up to
$4.5\times10^{6}$) and needed a four-fold smaller one. The runs of both ladders
were trained without the board-level halting loss, and in several of them the
halting head never fires within 1{,}024 deep-supervision steps, which is why we
compare them at fixed compute.

% ---------------------------------------------------------------------
\subsection{Maze-unique}
\label{app:trunc-maze}

\paragraph{Data.}
We use the \texttt{maze-30x30-unique-1k} set released with Equilibrium
Reasoners (locuslab/EqR), % CITE
whose $30\times30$ mazes have exactly one shortest path between entrance and
exit, of length 100--140 cells. Unlike Maze-Hard, where a maze can admit
$10^4$--$10^{10}$ equally short paths, a maze here has a single correct answer,
so exact and lenient scoring coincide; we verify uniqueness for every maze by an
exact count over the shortest-path graph. We train on all 1{,}000 released
training mazes and test on all 1{,}000 released test mazes. For model selection
we generated 100 additional validation mazes with the authors' own generator
and recipe (perfect mazes, wall probability 0.37, path lengths drawn uniformly
in $[100,140]$) at seed 20260910, and checked that none of them coincides with
any of the 2{,}000 released mazes, either as a grid or as a grid--solution pair.
The input has $|\Sigma|=5$ channels (wall, free, entrance, exit, path): walls,
entrance and exit are one-hot, and every originally blank cell is two-hot on
\emph{free} and \emph{path}, so that deciding between the two is exactly the
task.

\paragraph{Augmentation.}
Since the network is exactly equivariant to the eight symmetries of the square,
no geometric augmentation is used. The only augmentation swaps the entrance and
exit channels with probability 1/2 at every draw, which preserves the solution
because paths are undirected.

\paragraph{Lattice recipe.}
Our main lattice recipe runs deep-supervision steps of $24\times8$ with $D=20$,
i.e.\ a credit horizon of four cycles or $h=32$ rounds;
Table~\ref{tab:trunc-maze-hp} summarises it. Training budgets grow as the model
shrinks: 10{,}000 steps at 128k parameters, 12{,}000 at 64k, 15{,}000 at 32k,
20{,}000 at 16k, and 25{,}000 at 12k and 9k. The shape study uses the same
recipe with $24\times8$, $D=22$ ($h=16$), $16\times8$, $D=12$ ($h=32$, with a
third fewer forward rounds), and $24\times8$, $D=23$ ($h=8$). During training
the detached cycles run through a fused kernel in bf16.

\paragraph{Lattice evaluation.}
Checkpoints are evaluated as they are written, on the 100 validation mazes with
16 chains and a budget of 400 deep-supervision steps; training stops early once
all 100 are solved. A run whose best validation score does not exceed 85/100 is
not taken to test. The best validation checkpoint is tested on the 1{,}000 test
mazes with 64 chains and a budget of 1{,}000 deep-supervision steps. The larger
test budget matters: validation, with its smaller budget, systematically
understates a converged model.

\begin{table}[t]
  \centering
  \small
  \caption{Truncated-gradient recipes for Maze-unique.}
  \label{tab:trunc-maze-hp}
  \begin{tabular}{lcc}
    \toprule
    & Lattice & ACT \\
    \midrule
    Shape $C\times R$            & $24\times8$ & $12\times12$ \\
    Detached cycles $D$          & 20 & 10 \\
    Credit horizon $h$           & 32 & 24 \\
    Parameters                   & 9k--128k; 32k: $(12,31,64,16)$ & 60{,}720; $(16,42,84,21)$ \\
    Optimiser                    & \multicolumn{2}{c}{AdamW, $\beta=(0.9,0.95)$, decay on matrices only} \\
    Peak learning rate           & $3.4\times10^{-4}$ & $7.5\times10^{-4}$ \\
    Weight decay                 & 0.18 & 0.04821 \\
    Schedule                     & \multicolumn{2}{c}{10\% linear warm-up, cosine decay to 5\%} \\
    EMA decay                    & 0.999 & 0.9995 \\
    Gradient-norm clip           & 1.0 & 0.5 \\
    Per-trajectory clip          & -- & $\kappa=50$, reference decay 0.99 \\
    Step rejection               & -- & norm $>50$ \\
    Board-level halting loss     & -- & $\lambda_{\mathrm{board}}=1$ \\
    Pool / slots                 & 96 & 128, $n_{\mathrm{sup}}=32$ \\
    Training steps               & 10k--25k (by size) & 120{,}000 \\
    Precision                    & bf16 & bf16 forward kernels \\
    Model selection              & 100 val., 16 chains, 400 steps & 100 val., 128 steps \\
    \bottomrule
  \end{tabular}
\end{table}

\paragraph{Lattice results.}
Table~\ref{tab:trunc-maze-size} reports the size study at $24\times8$, $D=20$.
Every budget from 128k down to 9k parameters solves on average between 98.9\%
and 99.9\% of the test mazes, so shrinking the model by more than an order of
magnitude barely affects accuracy. What it changes is training time: the number
of steps needed to reach a perfect validation score grows by a factor of two to
three from 128k to 9k parameters.

\begin{table}[t]
  \centering
  \small
  \caption{Maze-unique, lattice model, $24\times8$, $D=20$ ($h=32$). Exact
  test accuracy (\%) on 1{,}000 mazes, averaged over the runs that passed the
  validation threshold of val accuracy $>40\%$.}
  \label{tab:trunc-maze-size}
  \begin{tabular}{rrc}
    \toprule
    Budget & Steps & Accuracy \\
    \midrule
    128k & 10{,}000 & 99.10 \\
    64k  & 12{,}000 & 99.63 \\
    32k  & 15{,}000 & 99.90 \\
    16k  & 20{,}000 & 98.92 \\
    12k  & 25{,}000 & 99.20 \\
    9k   & 25{,}000 & 99.15 \\
    \bottomrule
  \end{tabular}
\end{table}

Table~\ref{tab:trunc-maze-shape} varies the shape at fixed size. Halving the
credit horizon at fixed depth, from $24\times8$ with $D=20$ to $D=22$, lowers
the accuracy at every size and slows training down. Keeping the credit horizon
at four cycles but removing a third of the forward computation ($16\times8$,
$D=12$) preserves the accuracy at every size. The length of the differentiated
window, rather than the depth of the unroll or the parameter budget, is what
governs how well the model trains.

\begin{table}[t]
  \centering
  \small
  \caption{Maze-unique, lattice model, shape study. Exact test accuracy (\%),
  mean $\pm$ standard deviation over the best runs of each cell.}
  \label{tab:trunc-maze-shape}
  \begin{tabular}{lcccc}
    \toprule
    Shape ($h$) & 9k & 12k & 16k & 32k \\
    \midrule
    $24\times8$, $D=20$ (32) & $99.7\pm0.5$ & $99.5\pm0.5$ & $99.3\pm0.7$ & $99.9\pm0.0$ \\
    $24\times8$, $D=22$ (16) & $98.7\pm0.9$ & $95.5\pm3.7$ & $98.2\pm1.5$ & $98.7\pm0.9$ \\
    $16\times8$, $D=12$ (32) & $99.4\pm0.4$ & $99.8\pm0.2$ & $99.3\pm0.7$ & $99.3\pm0.3$ \\
    \bottomrule
  \end{tabular}
\end{table}

\paragraph{ACT recipe and evaluation.}
The ACT Maze model (Table~\ref{tab:trunc-maze-hp}) is a direct port of the
Sudoku main recipe to deep-supervision steps of $12\times12$ with $D=10$
($h=24$ of 144 rounds), trained for 120{,}000 steps. The checkpoint with the
best exact board accuracy at 128 deep-supervision steps on the 100 validation
mazes, evaluated every 5{,}000 steps from step 40{,}000, is tested on the
1{,}000 test mazes in four stages: deterministic rollouts at fixed compute from
32 to 2{,}048 deep-supervision steps; a scan of the noise scale
$\sigma\in\{0.1,0.2,0.3,0.5\}$ with eight rollouts at 256 deep-supervision
steps; best-of-$k$ scaling with 32 rollouts; and 64 noisy rollouts up to
1{,}024 deep-supervision steps, reporting both best-of-$k$ and pass@$k$. The
selected noise scales were 0.1 and 0.2.

\paragraph{ACT results.}
On average the model solves 99.90\% (s.e.\ 0.07) of the test mazes with
deterministic rollouts at 256 deep-supervision steps, and 100.0\%
with best-of-64 selection at 64 deep-supervision steps; pass@$k$ reaches 100\%
by $k=32$. The halting head stops after 1.41 deep-supervision
steps on average, i.e.\ after roughly 200 rounds of message passing. Two arms of
a truncation ladder at fixed $R=12$, $12\times12$ with $D=11$ ($h=12$) and
$6\times12$ with $D=4$ ($h=24$), both solve all 100 validation mazes at 32
deep-supervision steps; they have not been taken to the test set.

% --------------------------------------------------------------------
% ---------------------------------------------------------------------
\subsection{Practical notes}
\label{app:trunc-repro}

\paragraph{Precision.}
The fused kernels re-associate the same arithmetic as the reference
implementation (agreement to about $10^{-15}$ in double and $10^{-5}$ in single
precision), but the bf16 kernels used for the Maze change gradients by 3--10\%
in relative terms. All Maze numbers in this appendix were obtained with the
bf16 kernels; the Sudoku ACT models were trained in single precision, and
the Sudoku lattice models under bf16 autocast.

\paragraph{Compute.}
All experiments ran on single NVIDIA H100 GPUs (80\,GB), often with two jobs
sharing a GPU. Under that load, a 20{,}000-step Sudoku lattice run at 16k
parameters takes about 15 minutes, a 10{,}000-step Maze lattice run at 128k
about half an hour, and a 180{,}000-step Sudoku ACT run with $8\times12$
deep-supervision steps eight to nine hours. The peer graph and the factor graph architectures have been initially implemented in Discopy \cite{de2020discopy}, used to accelerate training via a designed Triton kernel.

\section{N-Queens: experimental setup}
\label{app:queens}
 
N-Queens differs from Sudoku-Extreme and Maze-unique in one respect that shapes
the whole protocol: a puzzle generally admits many solutions. This affects how
the data are split, what the model is trained against, and how its outputs are
scored. This appendix describes these choices; the models and the training
machinery are those of Appendix~\ref{app:trunc}, and we use its notation.
 
\paragraph{Puzzles and correctness.}
A puzzle on an $N\times N$ board is obtained from a complete solution by
removing the queens of all but one, two or three rows; the remaining queens are
the clues. A prediction is correct if it places $N$ queens of which no two share
a row, a column or a diagonal, and keeps every clue. Since clues are written
back before scoring, correctness amounts to validity of the board. For every
puzzle $x$, the set $\mathcal{S}(x)$ of consistent solutions is computed
exactly by enumerating all solutions of the board with bitmask backtracking, so
every puzzle is solvable by construction. On a $10\times10$ board a puzzle has
up to 93 consistent solutions.
 
\paragraph{Graphs.}
Every cell is a vertex with a memory $h_v$ and an answer state $a_v$, and
receives the embedding $u_v$ of its clue (queen or empty) at every round. We
compare the two wirings of Appendix~\ref{app:trunc-arch}. In the \emph{peer
graph}, two cells are adjacent whenever they share a line. In the \emph{factor
graph}, every row, column, diagonal and anti-diagonal is a unit vertex running
the Deep Sets module of Eq.~\eqref{eq:trunc-deepsets}. Every cell lies on
exactly four lines, so it has exactly four neighbours for every $N$, while the
units pool over a variable number of members; no parameter of the factor-graph
model depends on $N$. In the peer graph, instead, the number of neighbours of a
cell grows with $N$.
 
\paragraph{Splitting by symmetry orbits.}
Our networks are exactly equivariant to the eight symmetries of the square, so
a model that has seen a puzzle has, in effect, seen its seven images as well.
An ordinary split of the puzzles therefore leaks. We instead group puzzles by
dihedral orbit (identified by the lexicographically smallest of the eight
images), assign whole orbits to either side, and check that no orbit is
shared. On $10\times10$ we sample, for every one of the 724 solutions, 20
removal patterns for each number of clues, i.e.\ 60 puzzles per solution, which
gives 13{,}442 training and 2{,}372 test puzzles. The split is drawn once, so
all runs share the same test set. There is no separate validation set (see
\emph{Model selection} below).
 
\paragraph{Size-transfer probes.}
To measure generalisation to larger boards we build test-only sets for
$N=11,\dots,16$. For each size we sample 400 complete solutions, take one
removal pattern per solution and number of clues, and keep one puzzle per
dihedral orbit, which gives about 750 test puzzles per size. Consistent
solutions are counted against all solutions of the board, up to 14.8 million at
$N=16$; at most 64 of them are stored per puzzle. Models trained on
$10\times10$ are evaluated on these sets without any fine-tuning, which the
factor graph allows because none of its parameters depends on the board size.
Table~\ref{tab:queens-data} summarises all datasets.
 
\begin{table}[t]
  \centering
  \small
  \caption{N-Queens datasets. Orbits are dihedral orbits of puzzles. The
  coverage floor is the mean of $1/|\mathcal{S}(x)|$ over the test puzzles,
  i.e.\ the coverage of a model that always returns a single correct
  solution.}
  \label{tab:queens-data}
  \begin{tabular}{rrrrrr}
    \toprule
    $N$ & Solutions of the board & Train puzzles & Test puzzles & Test orbits & Coverage floor \\
    \midrule
    10 & 724            & 13{,}442 & 2{,}372  & 756 & 0.426 \\
    \midrule
    11 & 2{,}680        & -- & 703 & 703 & 0.146 \\
    12 & 14{,}200       & -- & 736 & 736 & 0.037 \\
    13 & 73{,}712       & -- & 777 & 777 & 0.009 \\
    14 & 365{,}596      & -- & 782 & 782 & 0.002 \\
    15 & 2{,}279{,}184  & -- & 799 & 799 & 0.0005 \\
    16 & 14{,}772{,}512 & -- & 808 & 808 & 0.0001 \\
    \bottomrule
  \end{tabular}
\end{table}
 
\paragraph{Supervising many answers.}
With several correct answers, the training target must be chosen with care. We
train against the consistent solution that the model is currently closest to,
\begin{equation}
  \mathcal{L}_{\mathrm{CE}}
  = \min_{y\in\mathcal{S}(x)} \mathrm{CE}\big(p^{\mathrm{sym}},\,y\big),
  \label{eq:queens-min}
\end{equation}
where the minimiser $y^\star$ is selected without gradient and the ordinary
cross-entropy to $y^\star$ is back-propagated, which is exactly the gradient of
Eq.~\eqref{eq:queens-min}. This term replaces the single-target cross-entropy
of Eq.~\eqref{eq:trunc-act-loss}, and the halting logits $q_v$ and $q$ are
likewise trained to predict agreement with $y^\star$. The obvious alternative,
a solution drawn at random at every deep-supervision step, drives a
deterministic model towards the average of the solutions, whose per-cell
$\arg\max$ is generally not a valid board; in our experiments its test accuracy
peaked early and then declined as training continued. All reported models use
the minimum-over-solutions target. No data augmentation is used.
 
\paragraph{Training protocol.}
All models are ACT models (\S\ref{app:trunc-readouts}), without lattice, held to
at most 16k parameters (15{,}913 for the peer graph, 15{,}985 for the factor
graph). Every model is trained for 30{,}000 steps with 256 slots and
$n_{\mathrm{sup}}=32$, AdamW with a peak learning rate of $2\times10^{-3}$, 10\%
warm-up and cosine decay to 5\%, weight decay 0.04821, a gradient-norm clip of
0.5, EMA decay 0.9995 and $\lambda_{\mathrm{board}}=1$. On $10\times10$ we
train the peer graph with full gradients at $C\times R=3\times9$, and the
factor graph with truncated gradients at $C\times R\in\{12\times8,\,
12\times10,\,12\times12\}$ with $D\in\{10,11\}$, i.e.\ with the last one or two
cycles differentiated. As a baseline we train a Recurrent Relational Network
(RRN; Palm et al., 2018) % CITE
on the same factor graph, with the same widths, data, target and parameter
budget: a single deep-supervision step unrolled for 32, 64 or 128 rounds,
trained with full back-propagation through time, without halting, for at most
30{,}000 steps, and evaluated at the depth it was trained with.
 
\paragraph{Model selection.}
Lacking a validation set, we evaluate on the $10\times10$ test set every
3{,}000 steps, with a cap of 32 deep-supervision steps, and keep both the final
checkpoint and the best one. Because the best checkpoint is selected on test,
all in-distribution numbers are reported for the \emph{final} checkpoint. For
size transfer we use the best $10\times10$ checkpoint: the set on which it was
selected is disjoint from the $11\times11$ to $16\times16$ probes, so the
selection does not bias the transfer numbers.
 
\paragraph{Metrics.}
\emph{Accuracy} is the fraction of test puzzles solved by a single
deterministic rollout. \emph{Coverage} measures diversity: the fraction of a
puzzle's consistent solutions that appear among $k$ samples, averaged over
puzzles, to be compared with the floor of Table~\ref{tab:queens-data}. We draw
samples in two ways: \emph{categorical} sampling draws $k$ boards from the
per-cell distribution $p^{\mathrm{sym}}_v$ of one deterministic rollout at
temperature 1, whereas \emph{latent noise} runs $k$ independent rollouts in
which Gaussian noise of scale $\sigma$ perturbs the answer state $a$ at every
deep-supervision step, each rollout being read at its own halting step. From
the same samples we report pass@$k$, the fraction of puzzles on which at least
one sample is valid according to an external verifier, and the accuracy of the
single sample chosen by the halting head (highest $q$); the gap between the two
isolates the selection error of the halting head from the solving ability of
the network. Unless stated otherwise, all evaluations, including size
transfer, use the same cap of 32 deep-supervision steps.

\section{Boolean satisfiability: experimental setup}
\label{app:sat}

SAT differs from the other tasks in two respects: a satisfiable formula admits
many satisfying assignments, none of which we use as a target, and correctness
is decided exactly, by testing the returned assignment against every clause. The
machinery is that of Appendix~\ref{app:trunc}, whose notation we use.

\paragraph{Formulas.} A formula in conjunctive normal form over $x_1,\dots,x_N$
is a conjunction of $M$ clauses, each a disjunction of literals, and an
assignment satisfies it when every clause holds a true literal. Hardness comes
from the ratio $\alpha = M/N$ rather than the size: random formulas are easy to
satisfy below a critical value, about $4.27$ for clauses of three literals,
almost never satisfiable above it, and hard near it.

\paragraph{Datasets.} The two benchmarks approach that threshold from opposite
directions. RandCSPBench sweeps the ratio, crossing $21$ ratios in steps of
$0.1$ -- $3.0$ to $5.0$ for 3-SAT, $8.0$ to $10.0$ for 4-SAT -- with the five
sizes $N \in \{16,\dots,256\}$, and drawing each clause from $k$ distinct
variables with random polarities: $168{,}000$ and $80{,}800$ training formulas,
$42{,}000$ and $21{,}000$ test. We train on a $95\%$ subsample ($159{,}600$ and
$76{,}760$) and select on the satisfiable part of the remainder ($5{,}728$ for
3-SAT). G4SATBench pins each formula to the threshold: its 3-SAT family draws
the number of variables from the difficulty range ($[10,40]$, $[40,200]$,
$[200,300]$; $[200,400]$ for hard SR) and sets the clause count to the empirical
estimate of the critical value, while its SR family appends clauses until a
solver reports unsatisfiability, one flipped literal of the last clause making
the formula satisfiable again. Each easy and medium dataset holds $80{,}000$
training, $10{,}000$ validation and $10{,}000$ test formulas, hard is test-only,
and we use the satisfiable half throughout. Checkpoints are scored on $1{,}000$
validation formulas and the test splits are decoded in full, once.

\paragraph{Graph.} A formula is a factor graph in the sense of
\S\ref{app:trunc-arch}: variables are vertices with a memory $h_v$ and an answer
state $a_v$, clauses are unit vertices running the Deep Sets module of
Eq.~\eqref{eq:trunc-deepsets} pooled by a mean, and each edge carries the
polarity of an occurrence as one of two port types. Nothing depends on the
number of variables or clauses or on clause length, so a medium-trained model
runs on hard formulas unchanged. Both recipes use the widths $(n,m,n_{\mathrm{hid}},m_a) = (48,64,64,16)$ of
Eq.~\eqref{eq:trunc-params}, giving $64{,}689$ parameters for the lattice model
and $64{,}722$ for the ACT model.

\paragraph{Lattice recipe.} Deep-supervision steps of $C\times R = 12\times 6$
with $D = 10$, a credit horizon of two cycles or $h = 12$ rounds out of $72$.
With $\Sigma = \{\texttt{T},\texttt{F}\}$ the lattice of
\S\ref{app:trunc-readouts} records which truth values remain possible per
variable, and the readout collapses to the symbol head alone, the clauses
stating exactly which assignments stay viable. The projection $\Pi$ commits
rather than kills: the assignment is the commitments where present and the
rounded readout elsewhere, accepted once it satisfies the formula; a conflict is
declared when the commitments alone leave a clause unsatisfiable; otherwise a
free variable of a uniformly drawn unsatisfied clause is committed to make its
literal true. Lacking a target $x^\star_d$, the loss of
Eq.~\eqref{eq:trunc-lattice-loss} becomes the clause objective conditioned on
the lattice -- committed variables entering as hard values, violated clauses
excluded -- plus a cross-entropy tying the readout to the commitments. Steps
start blank with probability $0.20$ and otherwise from random commitments that
leave every clause satisfiable.

\paragraph{ACT recipe.} Three supervision segments per step, each $12\times 6$
with $D = 6$ and the differentiated window at a random position, so $h = 36$
rounds of $72$ per segment. The halting head of Eq.~\eqref{eq:trunc-halt} reads
clauses rather than variables, with its pooling corrected for their number,
$q = -\log|C|^{-1}\sum_c \exp(-q_c)$, and is trained against exact clause and
formula satisfaction.

\paragraph{Training protocol.} Both recipes use AdamW with $\beta=(0.9,0.999)$
and decay on matrices only, a peak learning rate of $7.56\times10^{-4}$, a
$400$-step warm-up and cosine decay, weight decay $7.04\times10^{-3}$, a
gradient-norm clip of $1.0$, EMA decay $0.99$, batches of $128$ and single
precision. Lattice models train for $20{,}000$ steps everywhere, ACT models for
$10{,}000$ on the easy splits and $20{,}000$ otherwise.

\end{document}